\documentclass[10pt,twocolumn,letterpaper]{article}

\usepackage[pagenumbers]{cvpr} 

\usepackage[dvipsnames]{xcolor}

\usepackage[accsupp]{axessibility} 
\usepackage{multirow}
\usepackage{graphicx}
\usepackage{colortbl}
\usepackage{xcolor}
\usepackage{ulem}
\usepackage{pifont}
\usepackage{algorithm}
\usepackage{listings}
\usepackage{lipsum}
\usepackage{colortbl}
\usepackage{amsthm}
\usepackage{bbding}
\newcommand{\cmark}{\ding{51}}
\newcommand{\xmark}{\ding{55}}
\definecolor{chatgptgreen}{RGB}{16,163,127}
\definecolor{OTpurple}{RGB}{134,123,203}
\definecolor{Cyan_Light}{RGB}{238,255,255}
\definecolor{Cyan_Dark}{RGB}{220,255,255}
\definecolor{Yellow_Light}{RGB}{255,252,235}
\DeclareFixedFont{\ttb}{T1}{txtt}{bx}{n}{6} 
\DeclareFixedFont{\ttm}{T1}{txtt}{m}{n}{6}  

\definecolor{codegreen}{rgb}{0,0.6,0}
\definecolor{codegray}{rgb}{0.5,0.5,0.5}
\definecolor{codepurple}{rgb}{0.58,0,0.82}
\definecolor{backcolour}{rgb}{0.95,0.95,0.92}

\lstdefinestyle{mystyle}{
    commentstyle=\color{codegreen},
    keywordstyle=\color{magenta},
    numberstyle=\tiny\color{codegray},
    stringstyle=\color{codepurple},
    basicstyle=\ttfamily\footnotesize,
    breakatwhitespace=false,         
    breaklines=true,                 
    captionpos=b,                    
    keepspaces=true,                 
    numbers=left,                    
    numbersep=5pt,                  
    showspaces=false,                
    showstringspaces=false,
    showtabs=false,                  
    tabsize=2
}

\definecolor{cvprblue}{rgb}{0.21,0.49,0.74}
\usepackage[pagebackref,breaklinks,colorlinks,citecolor=cvprblue]{hyperref}

\def\paperID{15438} 
\def\confName{CVPR}
\def\confYear{2024}

\title{OST: Refining Text Knowledge with \uline{O}ptimal \uline{S}patio-\uline{T}emporal \\ Descriptor for General Video Recognition}
\author{Tongjia Chen$^1$, Hongshan Yu$^{1}\textsuperscript{\Envelope}$,  Zhengeng Yang$^{2}$, Zechuan Li$^1$, Wei Sun$^1$, Chen Chen$^{3}$ \\
$^1$Hunan University, $^2$Hunan Normal University \\
$^3$Center for Research in Computer Vision, University of Central Florida \\
{\small \Envelope~Corresponding author \qquad  \href{https://tomchen-ctj.github.io/OST}{Project Page: https://tomchen-ctj.github.io/OST}.} \\
}

\begin{document}
\maketitle
\begin{abstract}
Due to the resource-intensive nature of training vision-language models on expansive video data, a majority of studies have centered on adapting pre-trained image-language models to the video domain. Dominant pipelines propose to tackle the visual discrepancies with additional temporal learners while overlooking the substantial discrepancy for web-scaled descriptive narratives and concise action category names, leading to less distinct semantic space and potential performance limitations. In this work, we prioritize the refinement of text knowledge to facilitate generalizable video recognition. To address the limitations of the less distinct semantic space of category names, we prompt a large language model (LLM) to augment action class names into Spatio-Temporal Descriptors thus bridging the textual discrepancy and serving as a knowledge base for general recognition. Moreover, to assign the best descriptors with different video instances, we propose Optimal Descriptor Solver, forming the video recognition problem as solving the optimal matching flow across frame-level representations and descriptors. Comprehensive evaluations in zero-shot, few-shot, and fully supervised video recognition highlight the effectiveness of our approach. Our best model achieves a state-of-the-art zero-shot accuracy of 75.1\% on Kinetics-600.
\end{abstract}
\vspace{-4mm}
\section{Introduction}
\begin{figure}[t]
  \centering
\includegraphics[width=1\linewidth]{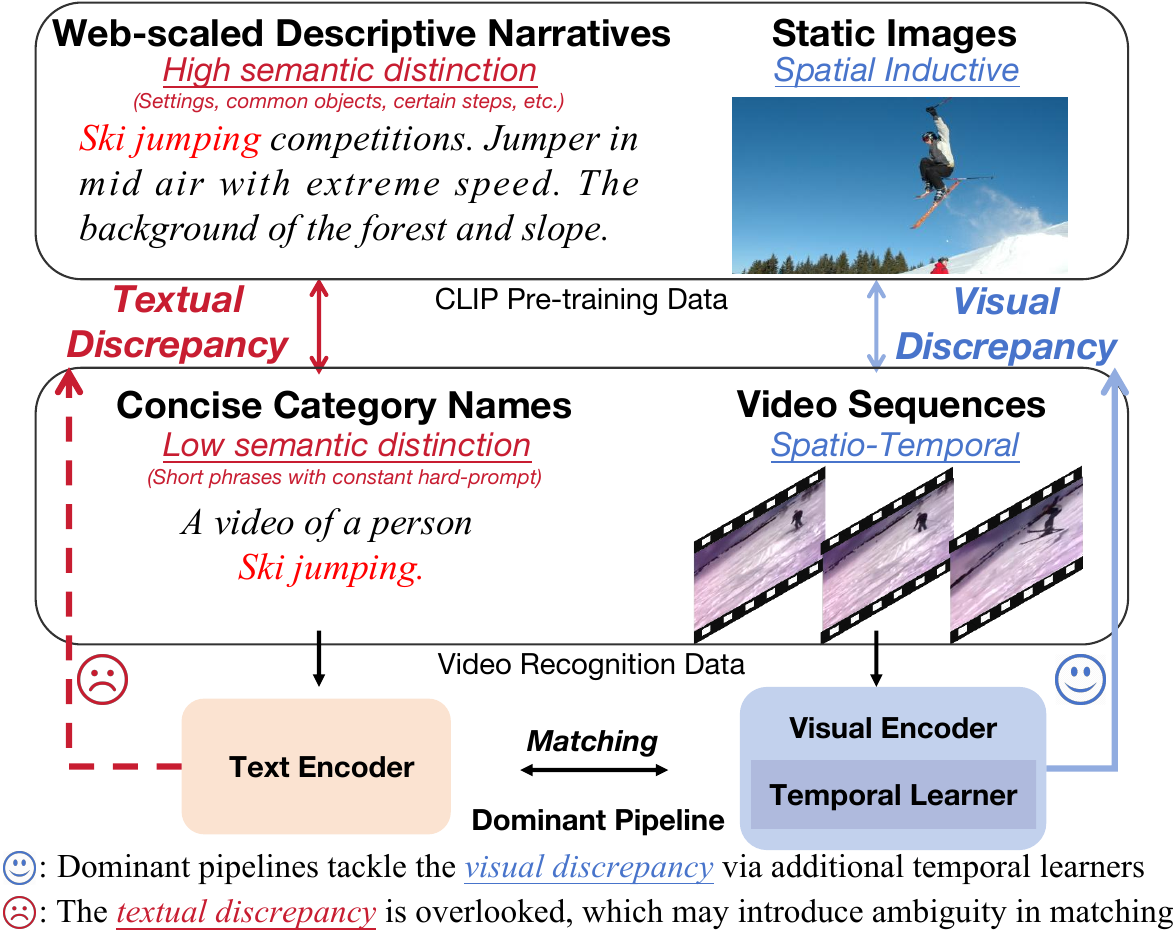}
   \vspace{-7mm}
   \caption{Motivation of our method. Dominant pipelines propose to tackle the visual discrepancies with additional temporal learners while overlooking the textual discrepancy between descriptive narratives and concise category names. This oversight results in a less separable latent space, which may hinder video recognition.}
   \label{fig: teaser}
   \vspace{-4mm}
\end{figure}
\label{sec:intro}
Large-scale contrastive language-image pre-training~\cite{CLIP, jia2021scaling,yuan2021florence} have shown remarkable performance in various computer vision tasks. The visual-semantic joint space not only serves powerful visual representation but also enables few/zero-shot transferring to downstream tasks with the reference of natural language. However, training a similar model for video recognition can be costly since large-scale video-language datasets are exponentially more massive~\cite{wang2023internvid} due to the extra temporal dimension. Hence, a feasible solution is to adapt the pre-trained image-text models for the task of video recognition. As depicted in Fig.~\ref{fig: teaser}, current methods devise a range of temporal learners to address the visual discrepancy while preserving text-domain knowledge in the semantic space of action category names, often by merging the category name with CLIP-style hard-prompts (\textit{e.g., ``a video of a person \{ski jumping\}"})~\cite{ACTIONCLIP, ILA, qing2023disentangling, text4vis, ni2022expanding}. Despite providing essential inter-class correlations that can benefit general recognition, we speculate this paradigm overlooks the textual discrepancy between web-scaled descriptive narratives in CLIP pre-training and concise category names in downstream video recognition. Given that category names of video datasets generally consist of verbs and nouns, the nouns exhibit variability while the verbs tend to remain consistent. For instance, \textit{playing cello, playing organ \& playing violin} are distinct actions related to playing instruments. The sole differentiation between these category names lies in the noun itself, resulting in low discriminative text embeddings. This may lead to a less separable semantic space, potentially introducing ambiguity in recognition tasks~\cite{bulat2023lasp}. 

\begin{figure}
  \centering
  \hfill
    \begin{subfigure}{0.47\linewidth}
    \includegraphics[width=1\linewidth]{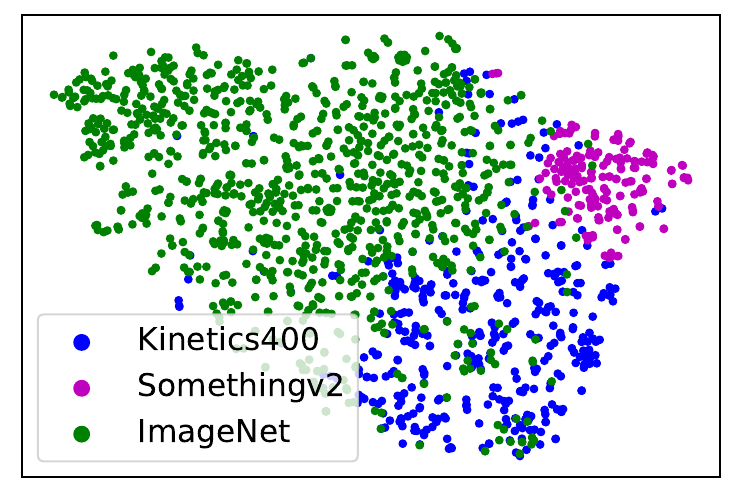}
    \label{fig:motivation-a}
    \vspace{-6mm}
  \end{subfigure}
  \centering
  \hfill
  \begin{subfigure}{0.47\linewidth}
    \includegraphics[width=1\linewidth]{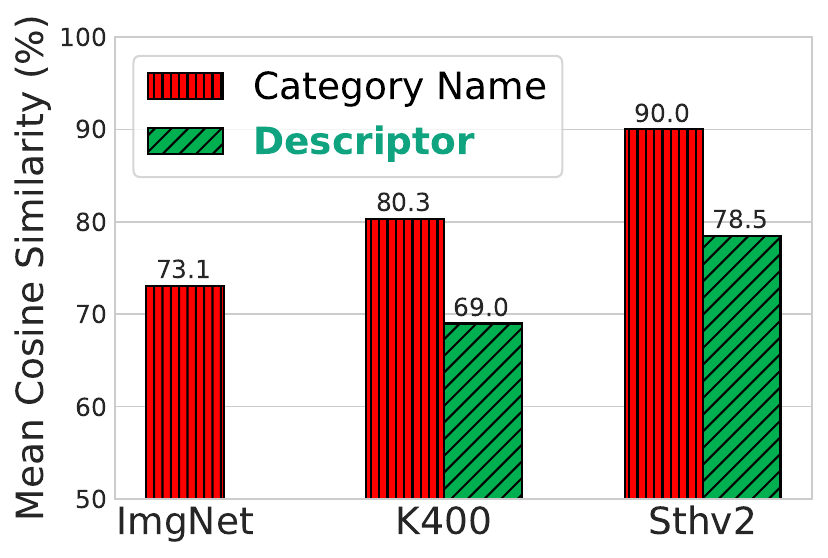}
    \label{fig:motivation-b}
    \vspace{-6mm}
  \end{subfigure}
  \caption{Sanity check on category names. We investigate the semantic distribution of video category names~\textbf{(Left)} and quantify the semantic density of category names~\textbf{(Right)}. We observe a higher semantic similarity of category names on K400 and Sthv2 compared to ImageNet. Our proposed \textit{Spatio-Temporal Descriptor} can greatly reduce the semantic similarity in latent space. \textit{Please refer to Sec.~\ref{sec: LLMAug} for comprehensive details.}}
  \label{fig: motivation}
  \vspace{-3mm}
\end{figure}

To validate our hypothesis, we perform a sanity check on the semantic distribution of category embeddings across ImageNet~\cite{deng2009imagenet}, Kinetics-400~\cite{quovadis}, and Something-Something v2~\cite{goyal2017something}. Initially, we employ a CLIP-B/16 text encoder to \uline{extract semantic embeddings of category names and leverage t-SNE visualization~\cite{tsne} to illustrate embedding clusters across the three datasets}. As depicted in Fig.~\ref{fig: motivation}~\textbf{(Left)}, features from K400 and Sthv2 datasets exhibit denser clustering compared to those from ImageNet, qualitatively indicating the low semantic distinction of video category names. To quantify this distinction and provide further support for our hypothesis, we \uline{compute pair-wise cosine similarity within each dataset and determine the average similarity}, serving as a measure of semantic density. A higher similarity implies a denser distribution of category embeddings and less separable semantics in the latent space. Fig.~\ref{fig: motivation}~\textbf{(Right)} visually demonstrates consistently higher mean cosine similarity of category names on video datasets compared to image datasets. This observation suggests that the intrinsic semantic space associated with video category names is less distinct. Since the category embedding serves as a decision plane~\cite{text4vis} in cross-modal matching (\textit{i.e.} compute the cosine similarity between category embeddings and visual features), such reduced distinctiveness may potentially diminish its efficacy in recognition tasks.

To mitigate this issue, one could manually craft textual narratives, but this process is labor-intensive. Alternatively, Large Language Models (LLMs) serve as a viable solution, acting as expansive knowledge bases that can generate detailed descriptors efficiently. As shown in Fig.~\ref{fig: teaser}, we can substantially refine our comprehension of ski jumping by integrating external contextual information such as the forest, the snow slope, and different action steps performed by the ski jumper. Hence, we propose to prompt LLMs with category names into what we define as \textit{Spatio-Temporal Descriptors} to enrich the semantic space with external knowledge. Where \textit{Spatio Descriptors} should possess the capability to capture static appearances, for instance, the environment and distinct objects included, while \textit{Temporal Descriptors} should focus on describing the temporal evolution of actions. This allows for the disentanglement of the category name into two complementary semantic spaces, thereby enhancing the semantic distinction and providing external knowledge for general recognition.

Based on the obtained descriptors, an intuitive solution is to aggregate these descriptors as a global category embedding via pooling, and match the embedding with corresponding visual features~\cite{kaul2023multi, menon2022visual}. However, this utilization might be suboptimal due to the following reasons: \textbf{1)}~Since the descriptors for one action class may not be contained in every video instance in this action category, directly matching the pooled descriptor-level representations with each video is potentially ineffective. \textbf{2)}~The propensity of LLMs to exhibit hallucinations~\cite{zhang2023siren} may bring noises to descriptors. To address this, we need to consider the adaptability of descriptors to individual video instances. 
In this vein, we propose \textit{Optimal Descriptor Solver} to obtain an optimal transport plan that adaptively aligns features across frame-level tokens and descriptors. 

In light of the above explorations, we propose \uline{O}ptimal \uline{S}patio-\uline{T}emporal Descriptor (\textbf{OST}), a general pipeline for video recognition. Our \textbf{OST} comprises two components: We first disentangle the category name into \textit{Spatio-Temporal Descriptors}, which not only bridges the semantic gap between narratives and category names but also serves as a knowledge base for general recognition. Then, we propose \textit{Optimal Descriptor Solver} that adaptively aligns frame-level representations with \textit{Spatio-Temporal Descriptors} to enhance video recognition. To demonstrate the effectiveness of our \textbf{OST}, we conduct comprehensive experiments on six benchmarks, including Kinetics-400~\cite{quovadis} \& 600~\cite{k600}, UCF-101~\cite{soomro2012ucf101}, HMDB-51~\cite{kuehne2011hmdb}, Something-Something V2~\cite{goyal2017something}, and ActivityNet~\cite{caba2015activitynet}. The results indicate that our method achieves state-of-the-art performance in open-vocabulary tasks, \textit{e.g.} zero-shot, few-shot, and also consistently improves the performance when combined with existing pipelines in fully-supervised settings. 
The main contributions of this work are as follows:
\begin{figure*}
  \centering
   \includegraphics[width=0.95\linewidth]{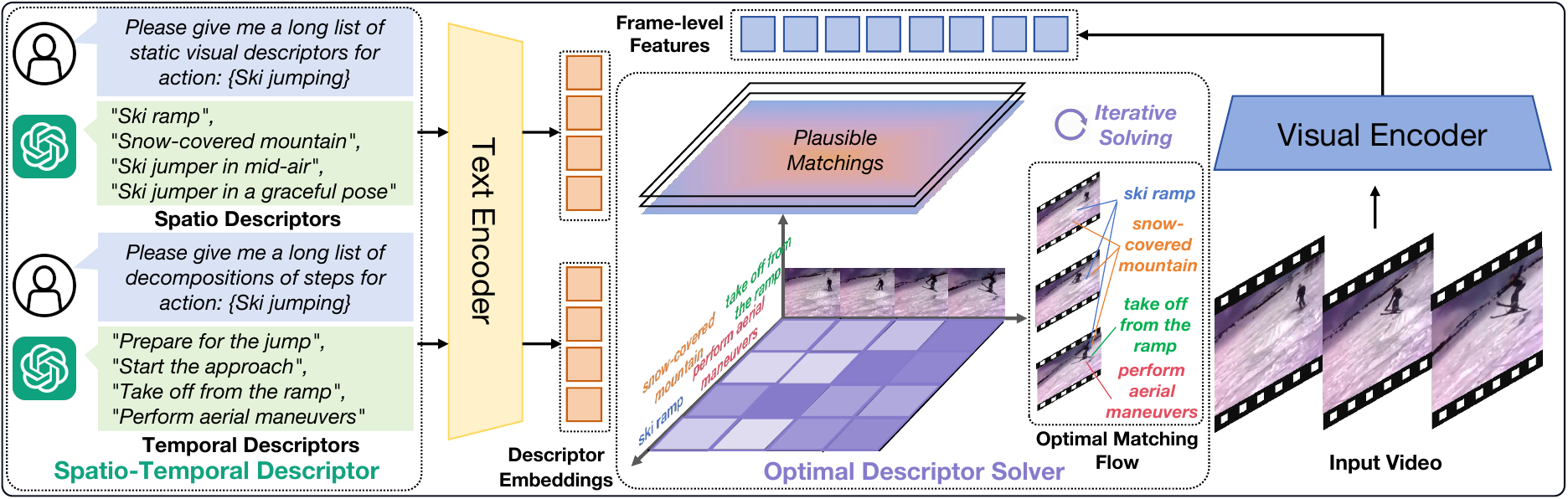}
   \vspace{-2mm}
   \caption{An overview of our pipeline for video recognition. We query the Large Language Model to augment category names to generate corresponding \textit{Category Descriptors}. The descriptors disentangled category names into \textit{Spatio-Temporal Descriptors} for static visual cues and temporal evolution, respectively. To fully refine the textual knowledge, we propose \textit{Optimal Descriptor Solver} that adaptively aligns descriptors with video frames. An optimal matching flow is calculated through the iterative solving of the entropy-regularized OT problem to assign optimal descriptors for each video instance. \textit{Please zoom in for comprehensive details.}}
   \label{fig: pipeline}
   \vspace{-5mm}
\end{figure*}

\begin{itemize}
\setlength\itemsep{-1.0em}
\item We provide new insights that prior pipelines for adapting vision-language pre-trained models to video recognition are constrained by the semantic space of category names. \\
\item We propose \textit{Spatio-Temporal Descriptors} derived from LLMs to enhance the distinction of semantic space and provide external knowledge for general recognition.\\
\item We introduce \textit{Optimal Descriptor Solver} that forms the video recognition problem as solving the optimal matching flow across frame-level representations and descriptors to fully refine the semantic knowledge. \\
\item Our \textbf{OST} presents a new way to utilize external knowledge to adapt pre-trained visual-language models for general video recognition. Experimental results in zero-shot, few-shot, and fully-supervised settings demonstrate the superior performance and generalizability of our method.
\end{itemize}
\section{Related Work}
\noindent\textbf{Video Recognition.} 
As a fundamental component of computer vision,
mainstream pipelines have typically explored traditional 2D, 3D CNNs~\cite{TSN, TSM, mutualnet, quovadis, R2P1D,hou2017tube} and Transformer-based methods~\cite{AIM, timesformer, videoswin, ni2022expanding, ILA, multiscaletransformer, EVL, chen2023first}. 
Additionally, methods modeling action phases~\cite{strafforello2023video, zhang2017learning, becattini2020done, zhou2023twinformer} have shown promise in video recognition, especially for long-form videos.
Recently, cross-modal video recognition~\cite{qing2023disentangling, ILA, ACTIONCLIP,xue2022clip,ju2022prompting,ni2022expanding,wu2023bike,text4vis} has benefited a lot from the powerful visual-text joint semantic space of CLIP. This cross-modal paradigm not only fosters strong representations with rich semantics but also achieves great open-vocabulary capacities. However, dominant pipelines~\cite{text4vis, ILA, qing2023disentangling, ACTIONCLIP} focus on the temporal discrepancies between images and videos while maintaining text-domain knowledge constantly. 
In contrast, our method prioritizes the refinement of text knowledge.

\noindent\textbf{Language for Visual Recognition.}
Differing from visual signals, natural language contains dense semantic information. Thus, language can serve as a rich source to provide inter-class correlations to benefit visual recognition. CuPL~\cite{cupl} and pipeline proposed by Menon \textit{et al.}~\cite{menon2022visual} utilizes category descriptions from GPT-3 as global category embedding for improved zero-shot image classification. Kaul \textit{et al.}~\cite{kaul2023multi} propose to utilize LLM descriptions and visual prototypes to construct a multi-modal classifier for enhanced open-vocabulary object detection. MAXI~\cite{lin2023match} proposes to construct text bags generated via multiple sources (\textit{e.g.,} captions and descriptions) to perform unsupervised finetuning for robust zero-shot action recognition. ASU~\cite{ASU} utilizes semantic units manually derived from WordNet and Wikipedia for video recognition. In this work, we aim to refine text knowledge by finding the optimal \textit{Spatio-Temporal Descriptors} automatically generated by LLMs to bridge the semantic discrepancy and provide external knowledge to benefit general video recognition.

\noindent\textbf{Optimal Transport.} 
Optimal Transport (OT), also known as Monge Problem~\cite{monge1781OT}, is an essential mathematical framework that facilitates the establishment of correspondences between two distinct distributions. Its great characteristics for distribution matching have benefited a variety of machine learning tasks~\cite{OTSurvey}, including domain adaptation~\cite{courty2017joint, damodaran2018deepjdot}, generative models~\cite{arjovsky2017wasserstein, gulrajani2017improved, salimans2018improving}, graph matching~\cite{chen2020graph, petric2019got}, image matching~\cite{Multiproxy, zhang2020deepemd}, and prompt learning~\cite{chen2023plot, kim2023zegot}, \textit{etc}. 
In this work, we propose to utilize OT distance to solve the cross-modal matching problem. To the best of our knowledge, this is the first work to form the video-text matching problem as solving the OT problem between frame-level representations and textual embeddings.
\vspace{-4mm}
\section{Method} 
In this section, we first review the preliminaries of optimal transport in Sec.~\ref{sec: Preliminaries}, then discuss our proposed \textit{Spatio-Temporal Descriptor} and \textit{Optimal Descriptor Solver} scheme in Sec.~\ref{sec: LLMAug} and Sec.~\ref{sec: ODSolver}, respectively. Finally, we introduce the training objectives in Sec.~\ref{sec: Objectives}.

\subsection{Preliminaries} \label{sec: Preliminaries}
Optimal transport aims to seek the minimal-cost transport plan between two distributions. In this work, we only consider the discrete distribution which is closely related to our framework. Assuming we have two sets of discrete empirical distributions:
\vspace{-4mm}
\begin{equation} \label{eq:1}
    \boldsymbol{\mu} = \sum^{M}_{i=1} p_{i} \delta_{x_{i}}, \quad \boldsymbol{\nu} = \sum^{N}_{j=1} q_{j} \delta_{y_{j}},
\vspace{-3mm}
\end{equation}
where $p_{i}$ and $q_{j}$ are the probability distribution summing to 1, $M$ and $N$ are number of samples in each empirical distribution, $\delta$ denotes the Dirac function. Since each certain distribution is discrete, the optimal transport plan $\boldsymbol{P}$ matching the two distributions is also discrete. In this setting, we can adapt Kantorovich OT formulation~\cite{kantorovich2006translocation} and form the optimal transport problem as:

\begin{equation} \label{eq:2}
\begin{aligned}
    & \boldsymbol{P}^{\ast} = \underset{\boldsymbol{P}\in \mathbb{R}^{M\times N}}{\arg{\min}} \sum^{M}_{i=1}\sum^{N}_{j=1} \boldsymbol{P}_{ij} \boldsymbol{C}_{ij}  \\
    & \textrm{s.t.} \quad \boldsymbol{P} \boldsymbol{e} = \boldsymbol{\mu}, \quad \boldsymbol{P}^{\top}\boldsymbol{e} = \boldsymbol{\nu}.        
\end{aligned}
\end{equation}
$\boldsymbol{C}\in \mathbb{R}^{M\times {N}}$ is the cost matrix that represents the distance between the support points $x_{i}$ and $y_{j}$ such as $\boldsymbol{C}_{ij} = 1 - sim(x_i, y_j)$. $\boldsymbol{P}^{\ast}$ is the optimal transport plan between two empirical distributions to minimize the total distance and $\boldsymbol{e}$ is the vector of ones.
Considering the computational and statistical limitations of this original OT formulation, we adopt the Sinkhorn-Knopp~\cite{cuturi2013sinkhorn} algorithm to solve the entropy-regularized OT problem. The regularized OT problem is defined as:
\begin{equation} \label{eq:3}
\begin{aligned}
    & \boldsymbol{P}^{\ast} = \underset{\boldsymbol{P}\in \mathbb{R}^{M\times N}}{\arg{\min}} \sum^{M}_{i=1}\sum^{N}_{j=1} \boldsymbol{P}_{ij} \boldsymbol{C}_{ij} 
    - \lambda\boldsymbol{H}(\boldsymbol{P})\\
    & \textrm{s.t.} \quad \boldsymbol{P} \boldsymbol{e} = \boldsymbol{\mu}, \quad \boldsymbol{P}^{\top}\boldsymbol{e} = \boldsymbol{\nu},        
\end{aligned}
\end{equation}
where $\boldsymbol{H}(\cdot)$ is the regularization operator and $\lambda$ is a regularization coefficient. Eq.~\ref{eq:3} is a convex problem and thus can be solved using the Sinkhorn algorithm. With $\boldsymbol{K} = \exp(-\boldsymbol{C}/\lambda)$, the regularized optimal transport can be computed by:
\begin{equation} \label{eq:4}
    \boldsymbol{P}^{\ast} = \text{diag}(\boldsymbol{a})\boldsymbol{K}\text{diag}(\boldsymbol{b}),
\end{equation}
where $\boldsymbol{a}$ and $\boldsymbol{b}$ are marginal constraints:
\begin{equation} \label{eq:5}
    \boldsymbol{a} \gets \boldsymbol{\mu}/ \boldsymbol{K}\boldsymbol{b}, \quad \boldsymbol{b} \gets \boldsymbol{\nu}/\boldsymbol{K}^{\top}\boldsymbol{a}.
\end{equation}

\subsection{Spatio-Temporal Descriptor} \label{sec: LLMAug}
In addressing the low semantic distinction of video categories, our objective is to disentangle category names into \textit{Spatio-Temporal Descriptors}. We posit that each type of descriptor yields information that is complementary to the other. \textit{Spatio Descriptors} are intended to capture static visual elements that can be discerned from a single image—such as settings and common objects. For \textit{Temporal Descriptors}, we aim to decompose the action classes in a step-by-step manner to describe the temporal evolution of an action. We use OpenAI's API for GPT-3.5~\cite{GPT} with a temperature of 0.7 to generate corresponding descriptors. 

To generate \textit{Spatio Descriptors}, inspired by~\cite{feng2023leveraging}, we use the following prompt $\mathcal{P}^s(\cdot)$ with category name $\boldsymbol{cls}$ to query LLM: ``\textit{Please give me a long list of descriptors for action: \{$\boldsymbol{cls}$\}, ${N_s}$ descriptors in total.}"\footnote{For a detailed demonstration of prompts we used, please refer to \textit{Supplementary Material}.}. This prompt enables the LLM to always return a list with ${N_s}$ descriptors. This process can be formulated as:
\begin{equation}
    \boldsymbol{Des^s} = \boldsymbol{LLM}[( \mathcal{P}^s(\boldsymbol{cls}))],
\end{equation}

For \textit{Temporal Descriptors}, we utilize the temporal prompt $\mathcal{P}^t(\cdot)$ as ``\textit{Please give me a long list of
decompositions of steps for action: \{$\boldsymbol{cls}$\}, ${N_t}$ steps in total}" and obtain ${N_t}$ descriptors:
\begin{equation}
    \boldsymbol{Des^t} = \boldsymbol{LLM}[( \mathcal{P}^t(\boldsymbol{cls}))].
\end{equation}
Nonetheless, our empirical study (\textit{please refer to Sec.\ref{sec: ablation}}) indicates that the direct application of temporal descriptors $\boldsymbol{Des^t}$, yields only marginal enhancements. As discussed in~\cite{momeni2023verbs, lin2023match, hendricks2021probing}, image-text pre-trained models are less sensitive to verbs. The initial semantic space of the temporal descriptors generated by CLIP might be limited. Thus, we adopt a hard prompt: ``\textit{A video of \{$\boldsymbol{cls}$\} usually includes \{$\boldsymbol{Des^t}$\}}" to condition temporal descriptors on the category names. We find this operation brings consistent improvements in recognition tasks.

Through this approach, we can disentangle the category name into two complementary semantic spaces. This disentanglement significantly mitigates the semantic similarity among class names and also serves sufficient knowledge for general recognition.

\begin{table*}[ht] \caption{Comparisons with state-of-the-art methods for zero-shot video recognition on HMDB51, UCF101 and Kinetics-600. We report Top-1 and Top-5 accuracy using single-view inference.}
\vspace{-2mm}
\label{table:zeroshot}
    \centering
    \setlength{\tabcolsep}{2.0pt}
    \scalebox{0.93}{
\begin{tabular}{lccccccc} 
\toprule
    \textbf{Method} & \textbf{Venue} & ~\textbf{Encoder}~ & ~\textbf{Frames}~ & ~\textbf{HMDB-51}~ & ~\textbf{UCF-101}~ & ~\textbf{K600 (Top-1)}~ & ~\textbf{K600 (Top-5)}~ \\
    \hline
 \rowcolor{gray!20}\multicolumn{8}{l}{\textit{Uni-modal zero-shot video recognition models}}\\ 
    ER-ZSAR~\cite{chen2021elaborative} & ICCV'21  & TSM      & 16    & 35.3 $\pm$ 4.6 & 51.8  $\pm$ 2.9  & 42.1 $\pm$ 1.4 & 73.1 $\pm$ 0.3 \\
    JigsawNet~\cite{qian2022rethinking}& ECCV'22  & R(2+1)D  & 16    & 38.7 $\pm$ 3.7 & 56.0  $\pm$ 3.1  &     -          &         -      \\
 \hline 
 \rowcolor{gray!20}\multicolumn{8}{l}{\textit{Adapting pre-trained CLIP}} \\ 
    Vanilla CLIP~\cite{CLIP}           & ICML'21  & ViT-B/16 & 32    & 40.8 $\pm$ 0.3 & 63.2  $\pm$ 0.2  & 59.8 $\pm$ 0.3 & 83.5 $\pm$ 0.2 \\
    ActionCLIP~\cite{ACTIONCLIP}       & arXiv'21 & ViT-B/16 & 32    & 40.8 $\pm$ 5.4 & 58.3  $\pm$ 3.4  & 66.7 $\pm$ 1.1 & 91.6 $\pm$ 0.3 \\
    Vita-CLIP~\cite{wasim2023vita}      & CVPR'23  & ViT-B/16 & 8~/~32& 48.6 $\pm$ 0.6 & 75.0  $\pm$ 0.6  & 67.4 $\pm$ 0.5 &       -        \\
    A5~\cite{ju2022prompting}          & ECCV'22  & ViT-B/16 & 32    & 44.3 $\pm$ 2.2 & 69.3  $\pm$ 4.2  & 55.8 $\pm$ 0.7 & 81.4 $\pm$ 0.3 \\
    XCLIP~\cite{ni2022expanding}       & ECCV'22  & ViT-B/16 & 32    & 44.6 $\pm$ 5.2 & 72.0  $\pm$ 2.3  & 65.2 $\pm$ 0.4 & 86.1 $\pm$ 0.8 \\
    DiST~\cite{qing2023disentangling}   & ICCV'23  & ViT-B/16 & 32    & \underline{55.4} $\pm$ 1.2 & 72.3  $\pm$ 0.6  &    -    &     -     \\
 \hline 
 \rowcolor{gray!20}\multicolumn{8}{l}{\textit{Tuning pre-trained CLIP}} \\     
    ViFi-CLIP~\cite{vificlip}          & CVPR'23  & ViT-B/16 & 32    & 51.3 $\pm$ 0.7 & 76.8  $\pm$ 0.8  & 71.2 $\pm$ 1.0 & 92.2 $\pm$ 0.3 \\
    MAXI~\cite{lin2023match}           & ICCV'23  & ViT-B/16 &16~/~32& 52.3 $\pm$ 0.6 & \underline{78.2}  $\pm$ 0.7  & \underline{71.5} $\pm$ 0.8 & \underline{92.5} $\pm$ 0.4 \\
  \hline 
  \multirow{2}{*}{\textbf{OST}} &   \multirow{2}{*}{CVPR'24}  &     \multirow{2}{*}{ViT-B/16}   &   8  & 54.9 $\pm$ 1.1 & 77.9  $\pm$ 1.3  & 73.9 $\pm$ 0.8 & 94.1 $\pm$ 0.3 \\ 
           &          &   & 32   & \cellcolor{green!20} \textbf{55.9} $\pm$ 1.2 & \cellcolor{green!20}\textbf{79.7} $\pm$ 1.1   &\cellcolor{green!20} \textbf{75.1} $\pm$ 0.6 & \cellcolor{green!20}\textbf{94.6} $\pm$ 0.2 \\
\bottomrule
\end{tabular}}
\vspace{-2mm}
\end{table*}
\subsection{Optimal Descriptor Solver} \label{sec: ODSolver}
A considerable number of transformer-based video recognition pipelines obtain video-level representation via pooling over image-level [CLS] tokens and then classify the video into a category by calculating the matching score using cosine similarity with category embeddings~\cite{vificlip, text4vis, ACTIONCLIP, ILA}, this pipeline can be formulated as:
\begin{equation} \label{eq:6}
    \boldsymbol{S}_k = cos(\boldsymbol{\overline{V}}, \boldsymbol{Cat_k}),
\end{equation}
where $cos(\cdot,\cdot)$ is the cosine similarity,  $\boldsymbol{V}\in \mathbb{R}^{T\times {d}}$ is a set of local representations with $T$ frames in total, $\boldsymbol{Cat_k} \in \mathbb{R}^{d}$ is category embedding for each class. As discussed before, only relying on the understanding of category names may lead to a less distinctive semantic space. After obtaining \textit{Spatio-Temporal Descriptors} introduced in Sec.~\ref{sec: LLMAug}, an intuitive operation is to form a global-level descriptor embedding to benefit visual recognition:
\begin{equation} \label{eq:7}
    \boldsymbol{S_k^s}_{pool} = cos(\boldsymbol{\overline{V}}, \boldsymbol{\overline{D_k^s}}),\quad \boldsymbol{S_k^t}_{pool} = cos(\boldsymbol{\overline{V}}, \boldsymbol{\overline{D_k^t}}),
\end{equation}
where $\boldsymbol{D_k}\in \mathbb{R}^{N\times{d}}$ is the embedding of \textit{Spatio-Temporal Descriptors}. By pooling along the $N$ dimension, we can obtain the discriminative global descriptor embedding. However, we find this formation can lead to sub-optimal performances: 1) By averaging the descriptor-level representations, the model treats all of the attributes equally. Since the descriptors are generated by an autoregressive language model without instance-level knowledge, these descriptors may not be contained in every video.  2) The hallucination problem of LLMs may bring noises to the descriptor. 

Hence, a natural question arises: \textit{\uline{how can we assign optimal descriptors for each video instance?}} In this regard, we introduce \textit{Optimal Descriptor Solver (OD Solver)}, by adapting optimal transport theory, we formulate the video-text matching problem as an optimal matching flow. After obtain a set of frame-level features $\boldsymbol{V}\in \mathbb{R}^{T\times {d}}$ and descriptor-level embedding for each class $\boldsymbol{D_k^s}\in \mathbb{R}^{N_s\times{d}}$, $\boldsymbol{D_k^t}\in \mathbb{R}^{N_t\times{d}}$. The cost matrix for each class can be defined as:
\begin{equation} \label{eq:8}
    \boldsymbol{C_k^s} = 1 - cos(\boldsymbol{V}, \boldsymbol{D_k^s}),\quad \boldsymbol{C_k^t} = 1 - cos(\boldsymbol{V}, \boldsymbol{D_k^t}).
\end{equation}
According to Eq.~\ref{eq:3}, the entropy-regularized OT problem can be defined as:
\begin{equation} \label{eq:9}
\begin{aligned}
    & \boldsymbol{P}^{\ast} = \underset{\boldsymbol{P}\in \mathbb{R}^{T\times N}}{\arg{\min}} \sum^{T}_{i=1}\sum^{N}_{j=1} \boldsymbol{P}_{ij} \boldsymbol{C}_{ij} 
    - \lambda\boldsymbol{H}(\boldsymbol{P})\\
    & \textrm{s.t.} \quad \boldsymbol{P} \boldsymbol{e} = \boldsymbol{\mu}, \quad \boldsymbol{P}^{\top}\boldsymbol{e} = \boldsymbol{\nu}.        
\end{aligned}
\vspace{-2mm}
\end{equation}

We can obtain the optimal transport plan $\boldsymbol{P_k^s}^{\ast}$ and $\boldsymbol{P_k^t}^{\ast}$ for \textit{Spatio-Temporal Descriptors} respectively by solving the convex problem in Eq.~\ref{eq:9} via the Sinkhorn algorithm as defined in Eq.~\ref{eq:4}. Here $\boldsymbol{P_k}^{\ast}\in \mathbb{R}^{T\times N}$ denotes the optimal matching flow between the video and descriptors.
The matching score based on the optimal matching flow can be obtained via Frobenius inner product:
\begin{equation} \label{eq:10}
\begin{aligned}
    &\boldsymbol{S_k^s}_{OT} = \sum^{T}_{i=1}\sum^{N}_{j=1} \boldsymbol{P_k^s}_{ij}^{\ast}cos(\boldsymbol{V}_{i}, \boldsymbol{D_k^s}_{j}),\\     &\boldsymbol{S_k^t}_{OT} = \sum^{T}_{i=1}\sum^{N}_{j=1} \boldsymbol{P_k^t}_{ij}^{\ast}cos(\boldsymbol{V}_{i}, \boldsymbol{D_k^t}_{j}).
\end{aligned}
\end{equation}

\begin{table*}[ht] \caption{Comparisons with state-of-the-art methods for few-shot video recognition on HMDB51, UCF101 and Something-Something V2. We scaled up the task to categorize all categories in the dataset with only a few samples per category for training. Here $K$ denotes training samples for each class. We report Top-1 accuracy using single-view inference.}
\vspace{-2mm}
 \label{table:fewshot}
    \centering
    \setlength{\tabcolsep}{2.0pt}
    \scalebox{0.93}{
\begin{tabular}{l llll|llll|llll}
  \toprule
  \multirow{2}{*}{\textbf{Method}}   & \multicolumn{4}{c}{\textbf{HMDB-51}} & \multicolumn{4}{c}{\textbf{UCF-101}} & \multicolumn{4}{c}{\textbf{SSv2}}\\
 \cmidrule(r){2-5}\cmidrule(r){6-9}\cmidrule(r){10-13}
  & $K$=$2$ & $K$=$4$ & $K$=$8$ & $K$=$16$ & $K$=$2$ & $K$=$4$ & $K$=$8$ & $K$=$16$ & $K$=$2$ & $K$=$4$ & $K$=$8$ & $K$=$16$\\
 \hline 
 \rowcolor{Cyan_Light}\multicolumn{13}{l}{\textit{Directly tuning on CLIP}} \\ 
Vanilla CLIP~\cite{CLIP}      & 41.9 & 41.9 & 41.9 & 41.9 & 63.6 & 63.6 & 63.6 & 63.6 & 2.7 & 2.7 & 2.7 & 2.7 \\
ActionCLIP~\cite{ACTIONCLIP}  & 47.5 & 57.9 & 57.3 & 59.1 & 70.6 & 71.5 & 73.0 & 91.4 & 4.1 & 5.8 & 8.4 & 11.1 \\
XCLIP~\cite{ni2022expanding}  & 53.0 & 57.3 & 62.8 & 64.0 & 48.5 & 75.6 & 83.7 & 91.4 & 3.9 & 4.5 & 6.8 & 10.0 \\
A5~\cite{ju2022prompting}     & 39.7 & 50.7 & 56.0 & 62.4 & 71.4 & 79.9 & 85.7 & 89.9 & 4.4 & 5.1 & 6.1 & 9.7 \\
ViFi-CLIP~\cite{vificlip}  & \underline{57.2} &\underline{62.7} & \underline{64.5}  & \underline{66.8}  &\underline{80.7}  &\underline{85.1} & \underline{90.0} & \underline{92.7}  &\underline{6.2}  &\underline{7.4}
& \underline{8.5} & \textbf{12.4} \\
\textbf{OST}~                      & \cellcolor{green!20}\textbf{59.1}\textsubscript{+1.9} & \cellcolor{green!20}\textbf{62.9}\textsubscript{+0.2} & \cellcolor{green!20}\textbf{64.9}\textsubscript{+0.4} & \cellcolor{green!20}\textbf{68.2}\textsubscript{+1.4} & \cellcolor{green!20}\textbf{82.5}\textsubscript{+1.8} & \cellcolor{green!20}\textbf{87.5}\textsubscript{+2.4} & \cellcolor{green!20}\textbf{91.7}\textsubscript{+1.7} & \cellcolor{green!20}\textbf{93.9}\textsubscript{+1.2} & \cellcolor{green!20}\textbf{7.0} \textsubscript{+0.8} & \cellcolor{green!20}\textbf{7.7} \textsubscript{+0.3} & \cellcolor{green!20}\textbf{8.9} \textsubscript{+0.4} &  \underline{12.2}  \\
\midrule
 \hline 
 \rowcolor{Yellow_Light}\multicolumn{13}{l}{\textit{Fine-tuned on K400}} \\ 
ViFi-CLIP~\cite{vificlip}  & 55.8 &\underline{60.5} & 64.3  &65.4  &84.0  &86.5 & 90.3 & 92.8  &6.6  &6.8 & 8.6 & 11.0 \\
MAXI~\cite{lin2023match}   & \underline{58.0} &60.1 & \underline{65.0}  &\underline{66.5}  &\underline{86.8}  &\underline{89.3} & \underline{92.4} & \underline{93.5}  & \underline{7.1}  &\underline{8.4} & \underline{9.3} & \underline{12.4} \\
\textbf{OST}~    &
\cellcolor{green!20}\textbf{64.8}\textsubscript{+6.8} & \cellcolor{green!20}\textbf{66.7}\textsubscript{+6.2} & \cellcolor{green!20}\textbf{69.2}\textsubscript{+4.2} & \cellcolor{green!20}\textbf{71.6}\textsubscript{+5.1} & \cellcolor{green!20}\textbf{90.3}\textsubscript{+3.5} & \cellcolor{green!20}\textbf{92.6}\textsubscript{+3.3} & \cellcolor{green!20}\textbf{94.4}\textsubscript{+2.0} & \cellcolor{green!20}\textbf{96.2}\textsubscript{+2.7} & \cellcolor{green!20}\textbf{8.0} \textsubscript{+0.9} & \cellcolor{green!20}\textbf{8.9} \textsubscript{+0.5} &  \cellcolor{green!20}\textbf{10.5}\textsubscript{+1.2} &
\cellcolor{green!20}\textbf{12.6}\textsubscript{+0.2} \\
\bottomrule
\end{tabular}}
\vspace{-2mm}
\end{table*}
By fusing the overall matching score in the Euclidean space and Wasserstein space described in Eq.~\ref{eq:7} and Eq.~\ref{eq:10} respectively, the overall logits can be expressed as:
\begin{equation} \label{eq:11}
    \boldsymbol{S_k}_{\textbf{OD}} = \frac{1}{4} (\boldsymbol{S_k^s}_{pool} + \boldsymbol{S_k^t}_{pool} + \boldsymbol{S_k^s}_{OT} + \boldsymbol{S_k^t}_{OT}).
\end{equation}
\textit{Please refer to Supplementary Material for pseudo-codes.}

\begin{table}[ht] \caption{Fully-supervised video recognition on Kinetics-400, Something-Something V2 and ActivityNet. We report Top-1 accuracy using single-view inference.}
\vspace{-2mm}
 \label{table:supervised}
    \centering
    \setlength{\tabcolsep}{2.0pt}
    \scalebox{0.93}{
\begin{tabular}{l llll}
  \toprule
  \multirow{2}{*}{\textbf{Method}}   & \multicolumn{4}{c}{\textbf{Encoder - Frames}} \\
 \cmidrule(r){2-5}
  & ~B/32~-~8~ & ~B/32~-~16~ & ~B/16~-~8~ & ~B/16~-~16~ \\
 \hline 
 \rowcolor{gray!20}\multicolumn{5}{l}{\textit{Kinetics-400}} \\ 
Text4Vis~\cite{text4vis}      & 78.5 & 79.3 & 81.4 & 82.6 \\
\textbf{OST}      & \textbf{78.7}({\color{chatgptgreen}\textbf{+0.2}}) & \textbf{79.8}({\color{chatgptgreen}\textbf{+0.5}}) & \textbf{82.0}({\color{chatgptgreen}\textbf{+0.6}}) & \textbf{83.2}({\color{chatgptgreen}\textbf{+0.6}}) \\
 \hline 
 \rowcolor{gray!20}\multicolumn{5}{l}{\textit{Something-Something V2}} \\ 
Text4Vis~\cite{text4vis}      & 54.3 & 56.1 & 57.9 & 59.9 \\
\textbf{OST}      & \textbf{54.4}({\color{chatgptgreen}\textbf{+0.1}}) & \textbf{56.4}({\color{chatgptgreen}\textbf{+0.3}}) & \textbf{58.4}({\color{chatgptgreen}\textbf{+0.5}}) & \textbf{60.3}({\color{chatgptgreen}\textbf{+0.4}}) \\
 \hline 
 \rowcolor{gray!20}\multicolumn{5}{l}{\textit{ActivityNet}} \\ 
Text4Vis~\cite{text4vis}      & 83.4 & 85.0 & 86.4 & 88.4 \\
\textbf{OST}      & \textbf{84.0}({\color{chatgptgreen}\textbf{+0.6}}) & \textbf{85.8}({\color{chatgptgreen}\textbf{+0.8}}) & \textbf{87.1}({\color{chatgptgreen}\textbf{+0.7}}) & \textbf{88.7}({\color{chatgptgreen}\textbf{+0.3}}) \\
\bottomrule
\end{tabular}}
\vspace{-6mm}
\end{table}
\subsection{Training Objectives} \label{sec: Objectives}
Considering the overall logits calculated by \textit{OD Solver} in Eq.~\ref{eq:11} can be described as video-to-text logits $\boldsymbol{S_{k}^{v2t}}_{\textbf{OD}} = \boldsymbol{OD}(\boldsymbol{V}, \boldsymbol{D^{s,t}_{k}})$. A symmetric text-to-video logits can be obtained via a similar way $\boldsymbol{S_{k}^{t2v}}_{\textbf{OD}} = \boldsymbol{OD}(\boldsymbol{D^{s,t}_{k}}, \boldsymbol{V})$. Then, the softmax-normalized similarity scores can be expressed as:
\begin{equation} \label{eq:12}
\begin{aligned}
    & \boldsymbol{p_{i}^{v2t}}_{\textbf{OD}}  = \frac{1}{K}\sum^{K}_{k=1}\frac{exp(\boldsymbol{S_{ki}^{v2t}}_{\textbf{OD}} / \tau)}{\sum^{B}_{j=1}exp(\boldsymbol{S_{kj}^{v2t}}_{\textbf{OD}} / \tau)}, \\
    & \boldsymbol{p_{i}^{t2v}}_{\textbf{OD}}  = \frac{1}{K}\sum^{K}_{k=1}\frac{exp(\boldsymbol{S_{ki}^{t2v}}_{\textbf{OD}} / \tau)}{\sum^{B}_{j=1}exp(\boldsymbol{S_{kj}^{t2v}}_{\textbf{OD}} / \tau)},
\end{aligned}
\end{equation}
where $\tau$ refers to the temperature hyperparameter for scaling, $B$ is the number of samples in the current mini-batch, and $K$ is the number of classes. Let $\boldsymbol{q^{v2t}}, \boldsymbol{q^{t2v}}$ denotes the ground-truth similarity scores, we can define the Kullback-Leibler (KL) divergence~\cite{kullback1951information} as the overall contrastive loss to optimize the model as:
\begin{equation} \label{eq:13}
    \mathcal{L}_{\textbf{OD}} = \frac{1}{2}[KL(\boldsymbol{p^{v2t}}_{\textbf{OD}}, \boldsymbol{q^{v2t}}) + KL(\boldsymbol{p^{t2v}}_{\textbf{OD}}, \boldsymbol{q^{t2v}})].
\end{equation}
\section{Experiments}
\noindent\textbf{Datasets.}
We conduct experiments across 6 video benchmarks: Kinetics-400~\cite{quovadis} \& 600~\cite{k600}, UCF-101~\cite{soomro2012ucf101}, HMDB-51~\cite{kuehne2011hmdb}, Something-Something V2~\cite{goyal2017something}, and ActivityNet~\cite{caba2015activitynet}. Our investigation encompasses various settings, including zero-shot, few-shot, and fully-supervised video recognition. \textit{See Supplementary Material for details.}

\noindent\textbf{Implementation Details.}
We employ a CLIP ViT-B/16 to conduct both zero-shot and few-shot experiments. We generate $N_{s,t} =4$ descriptors for each category. Following~\cite{lin2023match, ilharco2022patching, wortsman2022robust}, we perform a linear weight-space ensembling between the original CLIP and the finetuned model with a ratio of $0.2$. \textit{See Supplementary Material for details.}

\subsection{Main Results}
\noindent\textbf{Zero-shot video recognition.}  
We present our zero-shot video recognition results and compare our approach with SOTAs in Table~\ref{table:zeroshot}. The model is first fine-tuned on the Kinetics400 dataset and evaluated directly on downstream datasets to ascertain its generalization capacity with respect to unseen classes. Our approach outperforms regular uni-modal zero-shot video recognition pipelines by a large margin as shown in the upper table. Moreover, we draw comparisons with methods that use K400 to adapt CLIP models for zero-shot recognition. Noteworthy among these are methods that integrate additional temporal learners~\cite{ju2022prompting, qing2023disentangling,ni2022expanding} or employ VL prompting techniques~\cite{ju2022prompting, wasim2023vita}. Contrary to these approaches, our pipeline leverages refined textual knowledge to boost video recognition without altering the underlying architecture. We observe consistent improvements in all datasets with respect to these methods. 
\begin{table*} \caption{Ablation studies. We utilize ViT-B/16 as the backbone and use 8 frames for training/validation unless otherwise specified. All of the performances are top-1 accuracy (\%) in the zero-shot setting using single-view inference and spatial size of $224\times224$.}
 \label{table:ablation}
	\centering
		\begin{subtable}[th]{0.53\textwidth}
		\centering
		\scalebox{0.93}{
			\begin{tabular}{llll}
			\toprule
			\textbf{Method} &  \textbf{HMDB} & \textbf{UCF} & \textbf{K600} \\ 
                \midrule
			Category Name~\cite{vificlip} & 50.9  & 75.5 & 70.8 \\
			\textbf{Descriptors}*  & 53.3 ({\color{chatgptgreen}\textbf{+2.4}}) & 76.6 ({\color{chatgptgreen}\textbf{+1.1}}) & 69.3 \\
			\textbf{OD Solver}   & \textbf{54.5} ({\color{OTpurple}\textbf{+3.6}}) & \textbf{77.9} ({\color{OTpurple}\textbf{+2.4}}) & \textbf{72.3} ({\color{OTpurple}\textbf{+1.5}})  \\
			\bottomrule
			\end{tabular}}
                \caption{Study on cross-modal matching mechanisms. Here we apply the number of descriptors $N_{s,t}=4$. * denotes pooling descriptors along with category names.}
                \label{table: sub: matching mechanisms}
            \end{subtable}
            \hspace{2mm}
            \begin{subtable} [th]{0.40\textwidth}
            \centering
            \scalebox{0.93}{
			\begin{tabular}{ccccc}
			\toprule
			\textbf{Spatio} &  \textbf{Temporal} & \textbf{HMDB} & \textbf{UCF} & \textbf{K600}\\ 
                \midrule
                \cmark & \xmark & 46.7 & 65.3 & 56.3  \\
                \xmark & \cmark & 53.1 & 77.5 & 71.6 \\
                \cmark & \cmark & \textbf{54.5} & \textbf{77.9} & \textbf{72.3}  \\
			\bottomrule
			\end{tabular}}
                \caption{The impact of different descriptors. Here \cmark~means applying corresponding Spatio/Temporal descriptors.}
                \label{table: sub: stdescriptor}
            \end{subtable}
            \hspace{1mm}
            \begin{subtable} [th]{0.24\textwidth}
            \centering
            \scalebox{0.93}{
			\begin{tabular}{ccccc}
			\toprule
			\textbf{$N$} &  \textbf{HMDB} & \textbf{UCF}  & \textbf{K600} \\ 
                \midrule
			2 & 53.8 & 77.3 & 72.1\\
                4 & \textbf{54.5} & \textbf{77.9} & 72.3 \\
                8 & 53.0 & 77.5 & \textbf{72.6}  \\
			\bottomrule
			\end{tabular}}
                \caption{Comparisons between different number of descriptors $N$.}
                \label{table: sub: numsofN}
            \end{subtable}
            \hspace{1mm}
            \begin{subtable} [th]{0.38\textwidth}
            \centering
            \scalebox{0.93}{
			\begin{tabular}{ccccc}
			\toprule
			\textbf{Spatio} &  \textbf{Temporal} & \textbf{HMDB} & \textbf{UCF} & \textbf{K600}\\ 
                \midrule
                \xmark & \xmark & 49.8 & 74.1 & 64.2  \\
                \cmark & \xmark & 53.5 & \textbf{79.0} & 71.8  \\
                \cmark & \cmark & 53.5 & 78.9 & 72.1  \\
                \xmark & \cmark & \textbf{54.5} & 77.9 & \textbf{72.3}  \\
			\bottomrule
			\end{tabular}}
                \caption{Study on category conditioning operation. \cmark~means conditioning corresponding descriptors on category names.}
                \label{table: sub: conditioning}
            \end{subtable}
            \hspace{1mm}
            \begin{subtable} [th]{0.31\textwidth}
            \centering
            \scalebox{0.93}{
			\begin{tabular}{cccc}
			\toprule
			\textbf{Ensemble} &  \textbf{HMDB} & \textbf{UCF} & \textbf{K600} \\ 
                \midrule
			\xmark & 55.4            & 80.1 & 72.9           \\
                \cmark & \textbf{55.9}  & 79.7  & \textbf{75.1}  \\
			\bottomrule
			\end{tabular}}
                \caption{The effects of weight-space ensembling. \cmark~means perform ensemble with a ratio of $0.2$. 32 frames are used during training/validation.}
                \label{table: sub: ensemble}
            \end{subtable}
    \vspace{-2mm}
\end{table*}

We further compare our method with other fully finetuning paradigms~\cite{vificlip,lin2023match}. Serving as a baseline to our method, ViFi-CLIP~\cite{vificlip} relies on the direct utilization of category names to fine-tune the CLIP model. Notably, utilizing only 8 frames for training and validation, our method demonstrates competitive performance, surpassing our baseline by a large margin. Upon scaling up the input frames to 32, our method consistently exhibits improvements across all datasets in comparison to prior SOTAs. Even against MAXI~\cite{lin2023match} which leverages more diverse textual knowledge, such as frame-level captions, our approach showcases superior accuracy with a 3.6\% improvement on HMDB, 1.5\% on UCF, and 3.6\% on K600. 

\noindent\textbf{Few-shot video recognition.}  
We demonstrate our method's learning capacity and generalizability under the challenging all-way few-shot regime. The Top-1 accuracy on three datasets is reported in Table~\ref{table:fewshot}. We conduct 
\begin{table}[ht] \caption{Additional cost analysis of our method, we report step latency during training, and throughput~(TP) during inference. We refer to Top-1 as zero-shot accuracy on Kinetics-600. }
\vspace{-1mm}
 \label{table:costanalysis}
    \centering
    \setlength{\tabcolsep}{2.0pt}
    \scalebox{0.93}{
\begin{tabular}{l ccc}
  \toprule
  \textbf{Method}  & \textbf{Top-1}~(\%)~ & ~\textbf{Latency}~(\textit{s})~  & \textbf{TP}~(\textit{video/s}) \\
  \hline 
ViFi-CLIP~\cite{vificlip}  & 71.2    & \textbf{0.40} (1.0$\times$) & \textbf{40.9} (1.00$\times$)  \\
\textbf{OST} & \textbf{75.1}     & 0.44 (1.1$\times$)  & 40.0 (0.98$\times$) \\
\bottomrule
\end{tabular}}
\vspace{-10mm}
\end{table}
experiments in two different aspects. We first conduct an experiment that directly tunes CLIP for few-shot recognition. Our method shows consistent improvement over our baseline~\cite{vificlip} on HMDB-51, UCF101, and even temporal-heavy dataset SSv2.

Following~\cite{lin2023match}, we adopt our best model in zero-shot settings to further verify our method's generalization capacity. As a comparison, ViFi-CLIP shows degraded performance in this fashion (\textit{e.g.} $K=4$ on UCF, $K=16$ on SSv2). In this regime, our method outperforms the unsupervised contrastive training framework MAXI~\cite{lin2023match} in different shot settings by an average of $\sim$5\% on HMDB, $\sim$3\% on UCF, and $\sim$1\% on SSv2. This indicates the generalizability of our pipeline in the extremely low-shot settings. 

\noindent\textbf{Fully-supervised video recognition.}  
We also conduct fully-supervised experiments on three large-scale video benchmarks Kinetics-400, Something-Something V2, and ActivityNet to validate the effectiveness of our method in supervised settings. Serving as a standard pipeline to adapt pre-trained vision-language models for supervised video recognition, we choose Text4Vis~\cite{text4vis} as our baseline and vary different encoders ViT-B/32, and ViT-B/16 with 8, and 16 frames, respectively. As shown in Table~\ref{table:supervised}, we find our method improves upon our corresponding baseline for all different architectures on all datasets. We can see that the performance on K400 and SSv2 is about 0.5\% higher than Text4Vis~\cite{text4vis}. For ActivityNet, the accuracy is even 0.8\% higher than our counterparts. 

\subsection{Ablation Studies} \label{sec: ablation}
We conduct ablation studies on zero-shot settings in Table~\ref{table:ablation} to investigate our \textbf{OST}'s learning capacity and generalizability in different instantiations.

\noindent\textbf{Different cross-modal matching mechanisms.} 
Table~\ref{table: sub: matching mechanisms} shows the effects of different cross-modal matching mechanisms. For a fair comparison, we start with a baseline that uses the category name during matching as~\cite{vificlip}. By simply aggregating the \textit{Descriptors} along with the category name via mean pooling, the accuracy on HMDB and UCF improved by 2.4\% and 1.1\%, respectively. However, on the K600 dataset, we observe a 1.5\% performance drop. This validates our hypothesis that the enhanced distinction brought by pooling operation can benefit downstream recognition, but might not be optimal. We then introduce our \textit{OD Solver} to solve the optimal matching flow, we find that our approach can further boost the performance on HMDB and UCF, and achieve a remarkable improvement of 1.5\% on the large-scale dataset K600. Notably, the categories in the K600 validation set are more complicated compared to HMDB and UCF. This validates our \textit{OD Solver}'s effectiveness, especially in complicated open-vocabulary settings. 

\begin{figure*}[t]
\begin{center}
\includegraphics[width=0.96\linewidth]{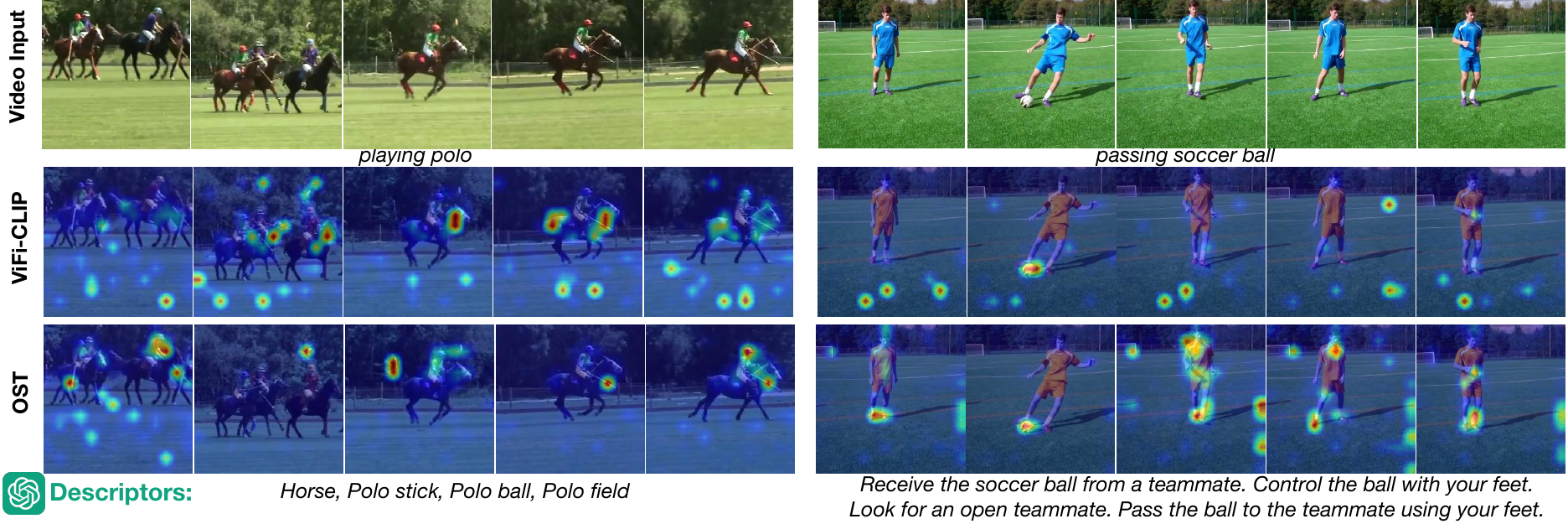}
\end{center}
\vspace{-7mm}
\caption{Attention map on K600 validation set. We demonstrate \textit{Spatio Descriptors} and \textit{Temporal Descriptors} on the left and right, respectively. \textbf{(Left):} For videos that can be recognized via static frames, our \textbf{OST} attends to the certain object more while ViFi-CLIP~\cite{vificlip} is often distracted by the backgrounds. \textbf{(Right):} For classes that require more temporal clues, ViFi-CLIP~\cite{vificlip} attends to appearance (\textit{e.g.} soccer ball and soccer field) more, while our \textbf{OST} shows consistent attention to the body's temporal salient parts such as the player's feet.}
\label{fig: attn_map_k600}
\vspace{-2mm}
\end{figure*}

\begin{figure*}[t]
\begin{center}
\includegraphics[width=0.96\linewidth]{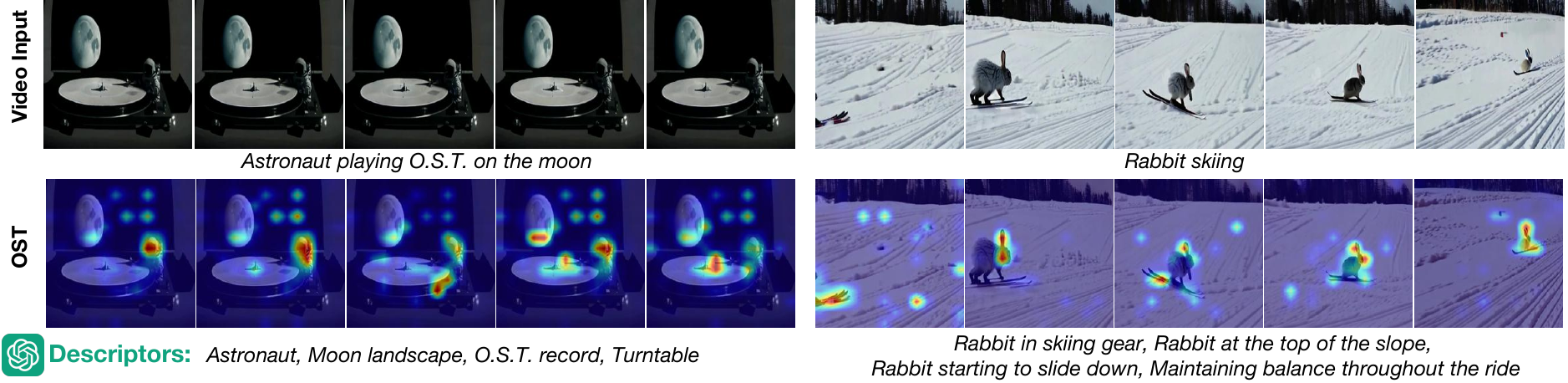}
\end{center}
\vspace{-7mm}
\caption{Generalization on extreme outliers. We utilize the text-to-video diffusion model Show-1~\cite{zhang2023show} to generate synthetic videos with a semantic distribution distinct from the fine-tuning data in Kinetics-400 to further demonstrate the generalizability of our method. Attention map for \textit{Spatio Descriptors} and \textit{Temporal Descriptors} are visualized on the left and right, respectively.}
\label{fig: extreme_outlier}
\vspace{-4mm}
\end{figure*}

\noindent\textbf{The impact of different descriptors.} 
We investigate the impact of \textit{Spatio-Temporal Descriptors} on the performance of our proposed method. The results shown in Table~\ref{table: sub: stdescriptor} demonstrate that each descriptor is complementary to others. Indicating that both \textit{Spatio} and \textit{Temporal Descriptors} provide crucial information for recognition tasks. We also observe that the effect of temporal descriptors is more convincing compared to \textit{Spatio Descriptors}. 

\noindent\textbf{Numbers of descriptors.}
We investigate the influences of varying the number of descriptors $N$ in Table~\ref{table: sub: numsofN}. We conducted experiments with 2, 4, and 8 \textit{Spatio-Temporal Descriptors}. We can observe that the performance reaches its peak at $N_{s,t}=4$. We've further checked the quality\footnote{Please refer to \textit{Supplementary Material} for examples of descriptors.} of descriptors when varying $N$. We find that 2 descriptors can not afford enough information to supply cross-modal matching. When the number of descriptors reaches 8, the hallucination problem of LLM becomes more severe, resulting in a significant amount of noisy descriptors. In this case, we set $N$ as 4 in our basic settings. 

\noindent\textbf{The impact of conditioning descriptors on category names.}
We study the effect of conditioning descriptors on category names on the final zero-shot accuracy. Table~\ref{table: sub: conditioning} shows that conditioning temporal descriptors on category names can achieve the best performances while conditioning both descriptors may lead to performance degradation. This further indicates the points framed in~\cite{momeni2023verbs, lin2023match, hendricks2021probing} that visual-language pre-trained models are less sensitive to verbs. As a result, the category conditioning technique can ensure the semantic distribution of the \textit{Temporal Descriptors} clustered well, making the optimization process smoother. 

\noindent\textbf{The effects of weight-space ensembling.}  
We investigate the effects of the linear weight-space ensembling technique. As shown in Table~\ref{table: sub: ensemble}, the ensembling technique greatly mitigates the catastrophic forgetting problem, especially on the large-scale Kinetics-600 dataset, where the zero-shot accuracy is improved by 2.2\%.
\vspace{-2mm}
\subsection{Cost Analysis}
\vspace{-2mm}
We analyze the additional cost of our method during training and inference in Table~\ref{table:costanalysis}. Latency is measured in our basic training setting and throughput is measured using the largest possible batch size before running out of memory with a single NVIDIA 4090-24G. Notably, the original implementation of ViFi-CLIP~\cite{vificlip} utilizes cross-entropy loss and maintains the logits for all categories in every mini-batch during training, leading to a larger latency. For a fair comparison, we re-implement ViFi-CLIP with local infoNCE-styled loss~\cite{ACTIONCLIP} to analyze the training cost. Our pipeline only requires an extra 0.1$\times$ training time and reduces the throughput by about 2\%, which is acceptable given the improvement in performance. 
\subsection{Visualizations}
We conduct a qualitative study on the attention map of our \textbf{OST} in the zero-shot setting. As depicted in Fig.~\ref{fig: attn_map_k600}, compared to our baseline ViFi-CLIP~\cite{vificlip} our method can not only focus on varied spatial cues but also consistently attend to temporal salient elements (\textit{e.g.} the player's feet) for videos that include more scene dynamics. Additionally, we investigate the attention map of our method on extreme outlier samples in Fig.~\ref{fig: extreme_outlier}. Our empirical findings indicate that out \textbf{OST} upholds robust generalization capabilities, even in extreme out-of-distribution examples. \textit{Please refer to Supplementary Material for more qualitative results.}
\vspace{-2mm}


\section{Conclusion}
In this work, we introduce a novel general video recognition pipeline \textbf{OST}. We prompt an LLM to augment category names into \textit{Spatio-Temporal Descriptors} and refine the semantic knowledge via \textit{Optimal Descriptor Solver}. Comprehensive evaluations in six datasets and three different tasks demonstrate the effectiveness of our approach. 
\vspace{-3mm}
\section*{Acknowledgement}
\vspace{-1.5mm}
The work was done while Tongjia was a research intern mentored by Chen Chen. We thank Ming Li and Yong He for proofreading and discussion. 
{
    \small
    \bibliographystyle{ieeenat_fullname}
    \bibliography{main}
}

\clearpage
\setcounter{page}{1}
\maketitlesupplementary

\section{Overview of Supplementary Material}\label{A}
In the supplementary material, we provide additional details in the following sections:\\
\begin{itemize}
\setlength\itemsep{0em}
\item Section~\ref{E}:~Further Analysis and Experiments\\
\item Section~\ref{B}:~Details of Optimal Descriptor Solver \\
\item Section~\ref{C}:~Dataset and Implementation Details\\
\item Section~\ref{D}:~Demonstration of Prompts and Descriptors\\
\item Section~\ref{H}:~Broader Impact and Limitation
\end{itemize}

\section{Further Analysis and Experiments}\label{E}
\subsection{Visualizations of Adaptive Transport Plan}
\begin{figure*}[ht]
\centering

\begin{subfigure}{\textwidth}
  \centering
  \includegraphics[width=.95\linewidth]{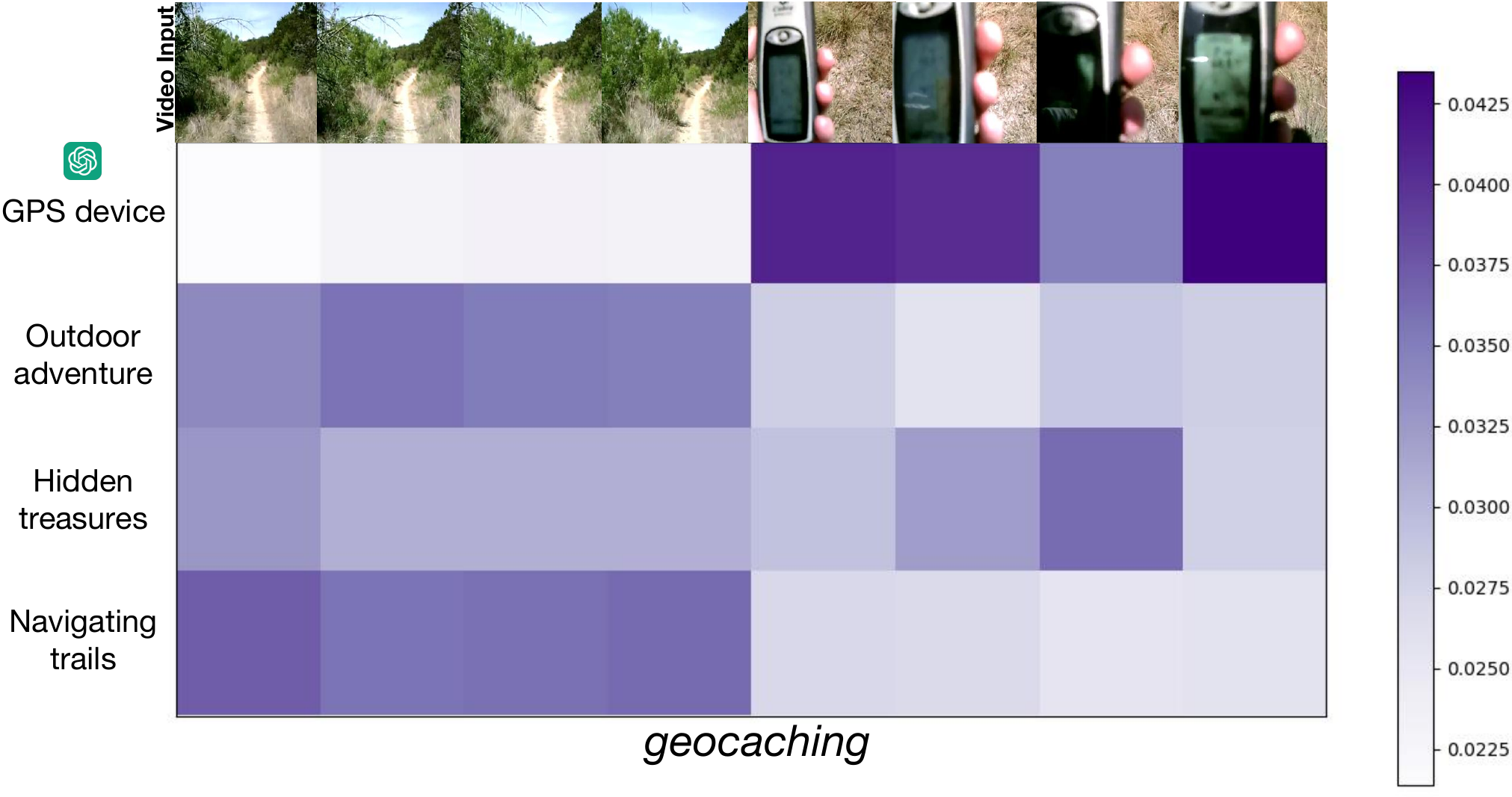}
  \caption{Adaptive transport plan of action category ``\textit{geocaching}''}
  \label{supp: tp_s_sub1}
\end{subfigure}%
\newline

\begin{subfigure}{\textwidth}
  \centering
  \includegraphics[width=.95\linewidth]{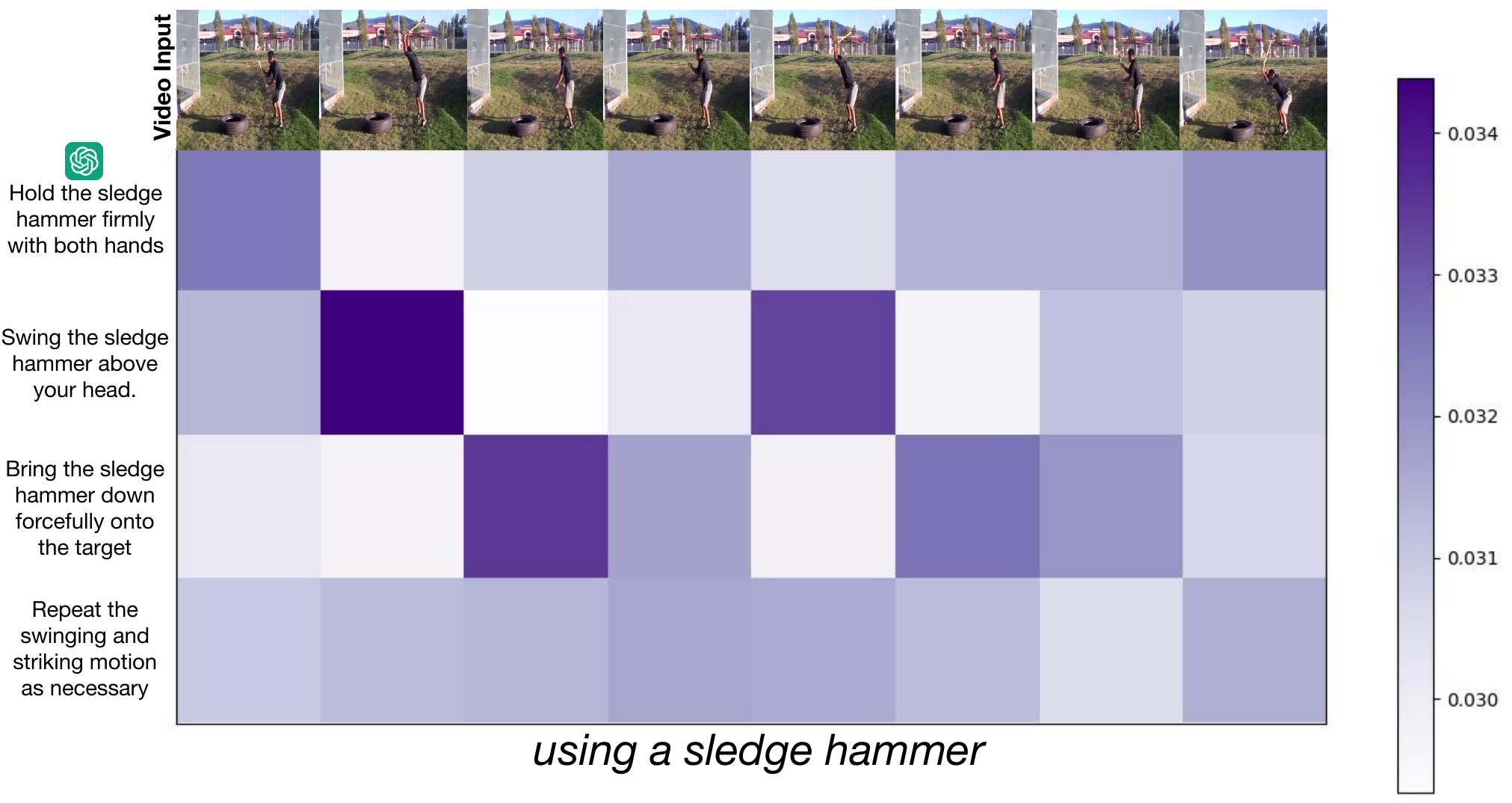}
  \caption{Adaptive transport plan of action category ``\textit{using a sledge hammer}''}
  \label{supp: tp_s_sub2}
\end{subfigure}

\caption{Visualization of the adaptive transport plan. Our \textit{OD Solver} not only integrates various visual cues—such as \textit{GPS devices}, \textit{navigation trails} in Fig.~\ref{supp: tp_s_sub1}, and \textit{hammer-swinging motions} in Fig.~\ref{supp: tp_s_sub2}, but also greatly reduce the detrimental effects of the noisy descriptors that often arise from the hallucination issues associated with LLMs, such as misleading `\textit{hidden treasures}' in Fig.~\ref{supp: tp_s_sub1} or `\textit{repeat the swinging}' in Fig.~\ref{supp: tp_s_sub2}). It is important to note that while the absolute variances among transport plans are relatively small, their substantial relative differences are critical in optimal matching.}
\label{supp: tp_s}
\end{figure*}

\begin{figure*}[ht]
\centering

\begin{subfigure}{\textwidth}
  \centering
  \includegraphics[width=.95\linewidth]{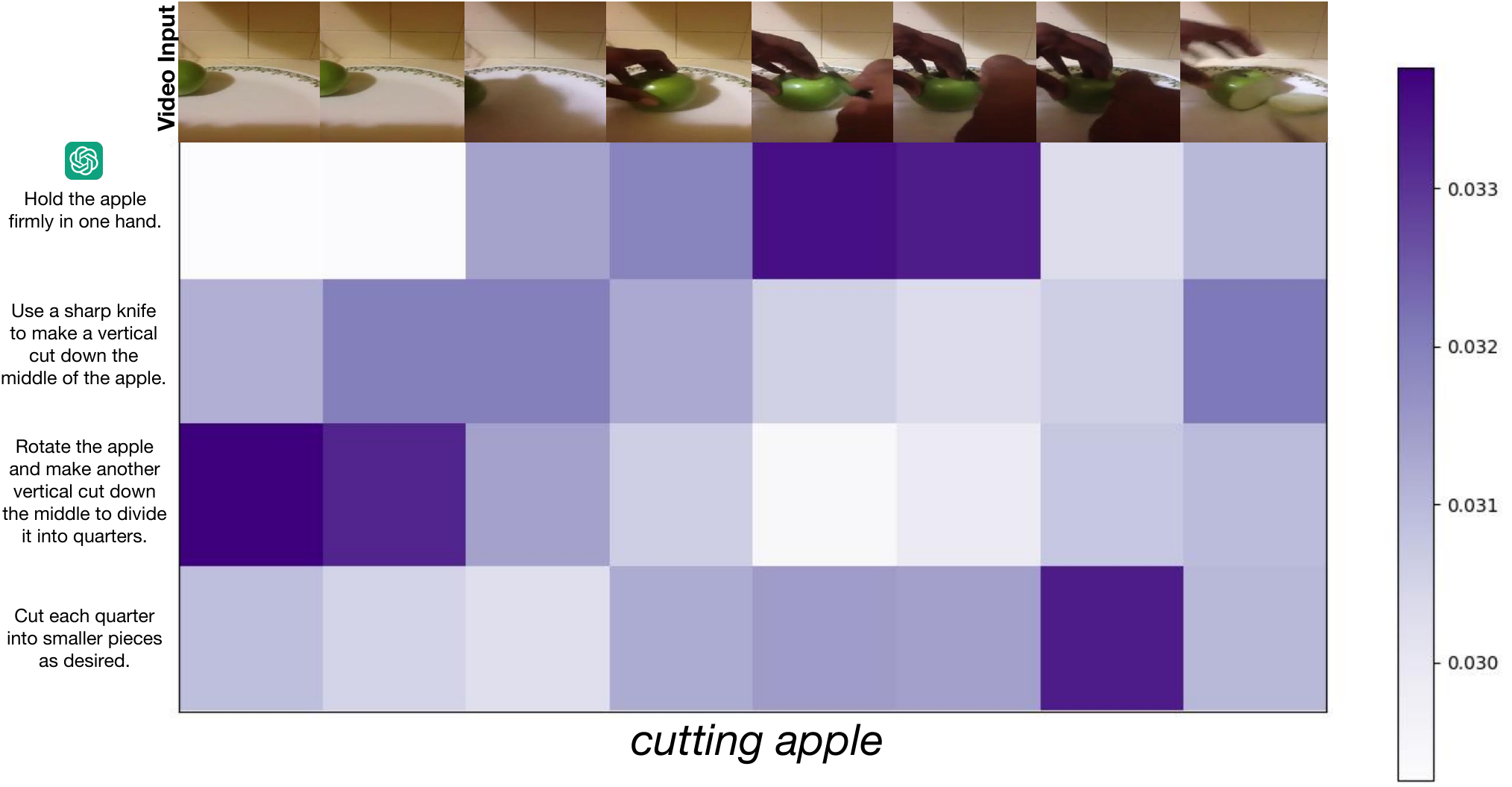}
  \caption{Adaptive transport plan of action category ``\textit{cutting apple}''}
  \label{supp: tp_t_sub1}
\end{subfigure}%
\newline

\begin{subfigure}{\textwidth}
  \centering
  \includegraphics[width=.95\linewidth]{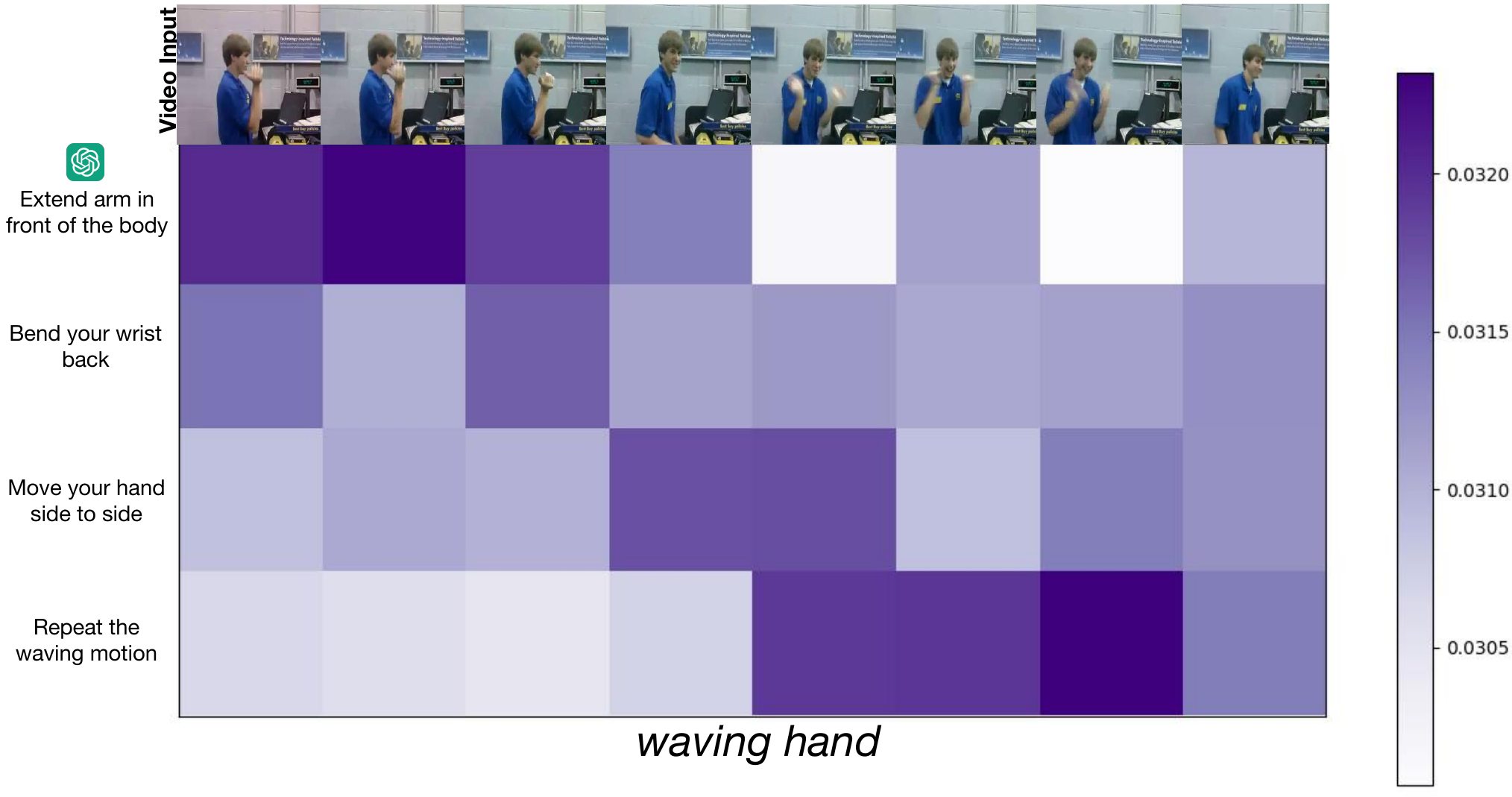}
  \caption{Adaptive transport plan of action category ``\textit{waving hand}''}
  \label{supp: tp_t_sub2}
\end{subfigure}

\caption{Visualization of the adaptive transport plan. Our investigation reveals that our \textit{OD Solver} can synchronize different action steps described by \textit{Temporal Descriptor} with corresponding video sequences. For example, it accurately coordinates actions such as `\textit{holding an apple firmly in one hand}' in Fig.~\ref{supp: tp_t_sub1}, `\textit{extend the arm in front of the body}', and `\textit{moving the hand from side to side}' in Fig.~\ref{supp: tp_t_sub2} with the help of corresponding \textit{Temporal Descriptors}.}
\label{supp: tp_t}
\end{figure*}

We analyze the adaptive transport plan in our proposed \textit{OD Solver}. Qualitative visualizations of the transport plan are illustrated in Fig.~\ref{supp: tp_s} and Fig.~\ref{supp: tp_t}, with detailed explanations provided in the captions. We find that our proposed \textit{OD Solver} can adaptly assign each descriptor to the video instance.
\subsection{Visualizations of Attention Maps}
\begin{figure*}
  \centering
   \includegraphics[width=0.96\linewidth]{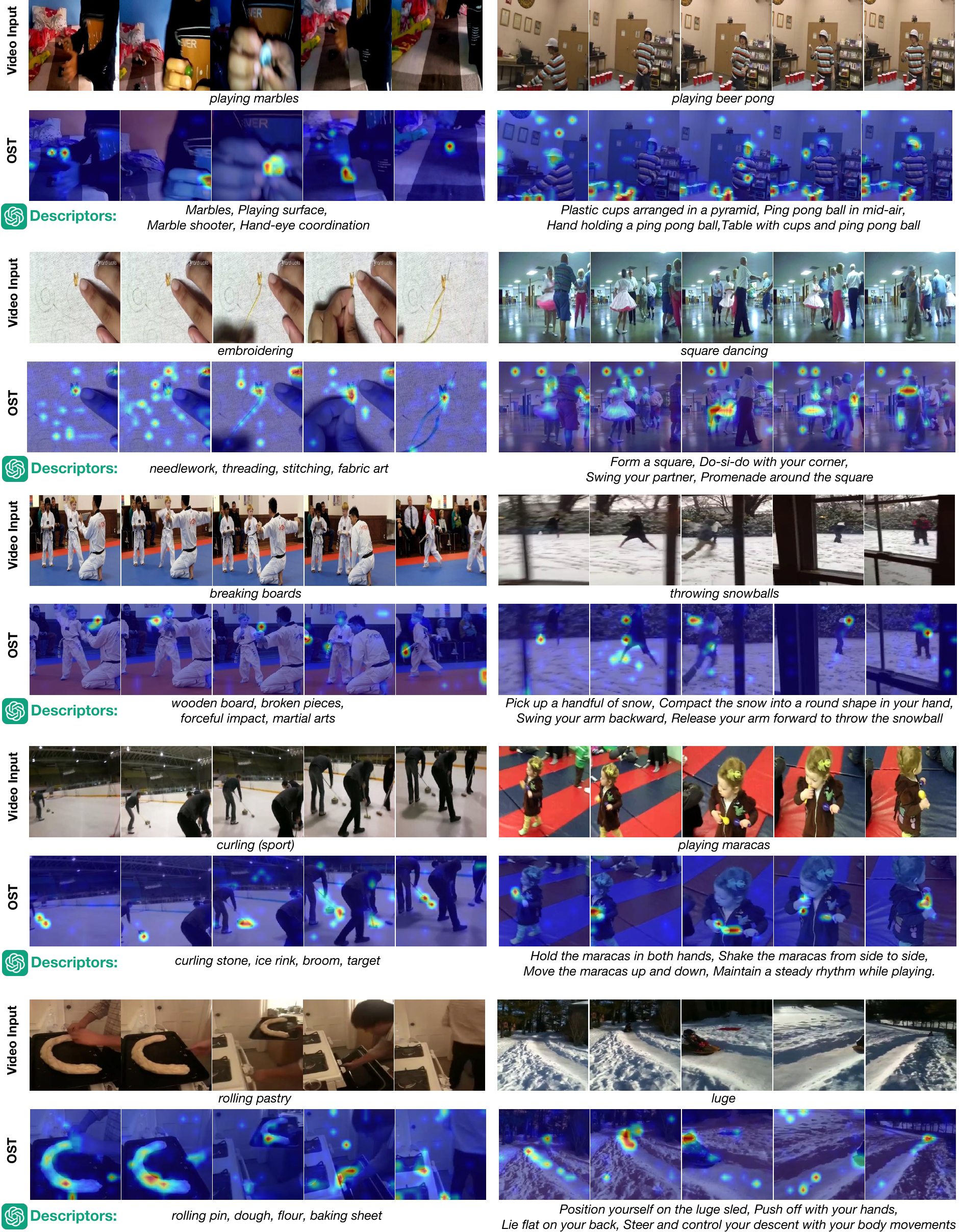}
   \caption{Additional visualizations of attention maps. The attention maps corresponding to the \textit{Spatio Descriptors} and \textit{Temporal Descriptors} are depicted on the left and right sides, respectively. The visualizations reveal that our proposed \textbf{OST} consistently focuses on specific static objects and temporal salient parts. This consistent focus underscores the efficacy of our approach.}
   \label{supp: attn_map}
\end{figure*}

We provide additional visualizations of the attention maps of our proposed \textbf{OST} in Fig.~\ref{supp: attn_map}.
\subsection{The Robustness of OST}
\begin{figure*}[ht]
\centering

\begin{subfigure}{\textwidth}
  \centering
  \includegraphics[width=.95\linewidth]{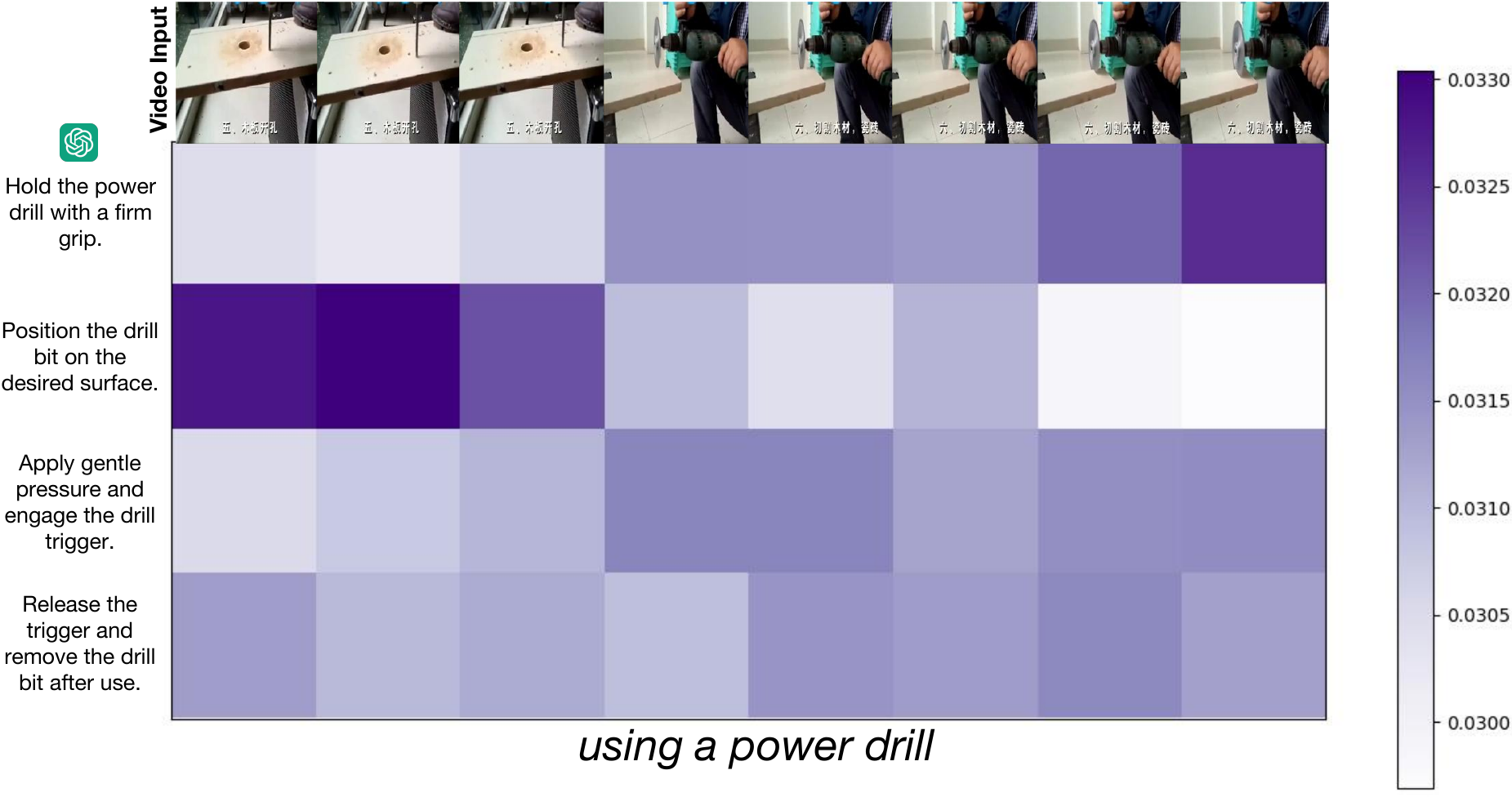}
  \caption{Adaptive transport plan of action category ``\textit{using a power drill}''}
  \label{supp: tp_t_sub3}
\end{subfigure}%
\newline

\begin{subfigure}{\textwidth}
  \centering
  \includegraphics[width=.95\linewidth]{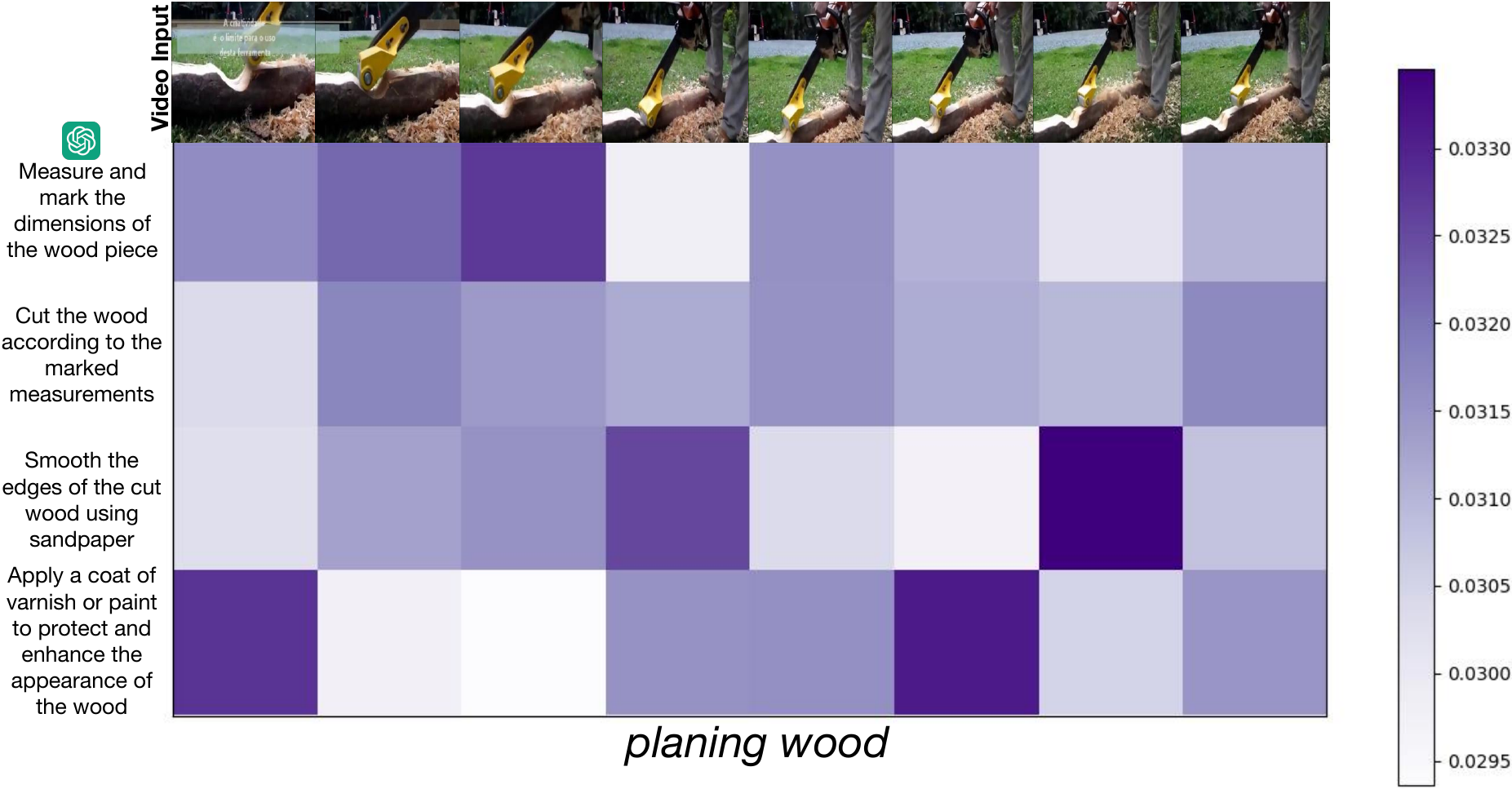}
  \caption{Adaptive transport plan of action category ``\textit{planing wood}''}
  \label{supp: tp_t_sub4}
\end{subfigure}

\caption{Visualization of cases where certain steps are missing. Our study demonstrates the efficacy of our proposed \textit{OD Solver} in accurately identifying instances when specific action steps are either missing or altered. As depicted in Fig.~\ref{supp: tp_t_sub3}, the actions `\textit{Engage the drill trigger}’ and `\textit{Release the trigger}’ are absent from the video sequence. With the help of our proposed \textit{OD Solver}, our model is capable of adaptively aligning the video instance with its corresponding category descriptor, effectively compensating for these absences. This capability is further evidenced in Fig.\ref{supp: tp_t_sub4}, where the action `\textit{Cut the wood according to the marked measurements}’ is missing from the video instance. Our \textit{OD Solver} adeptly adjusts to the modified sequence by assigning lower weights to the descriptors associated with the missing actions, demonstrating the method's robustness in handling incomplete or altered action sequences.}

\label{supp: tp_t_rebuttal}
\end{figure*}

\begin{figure*}
  \centering
   \includegraphics[width=0.96\linewidth]{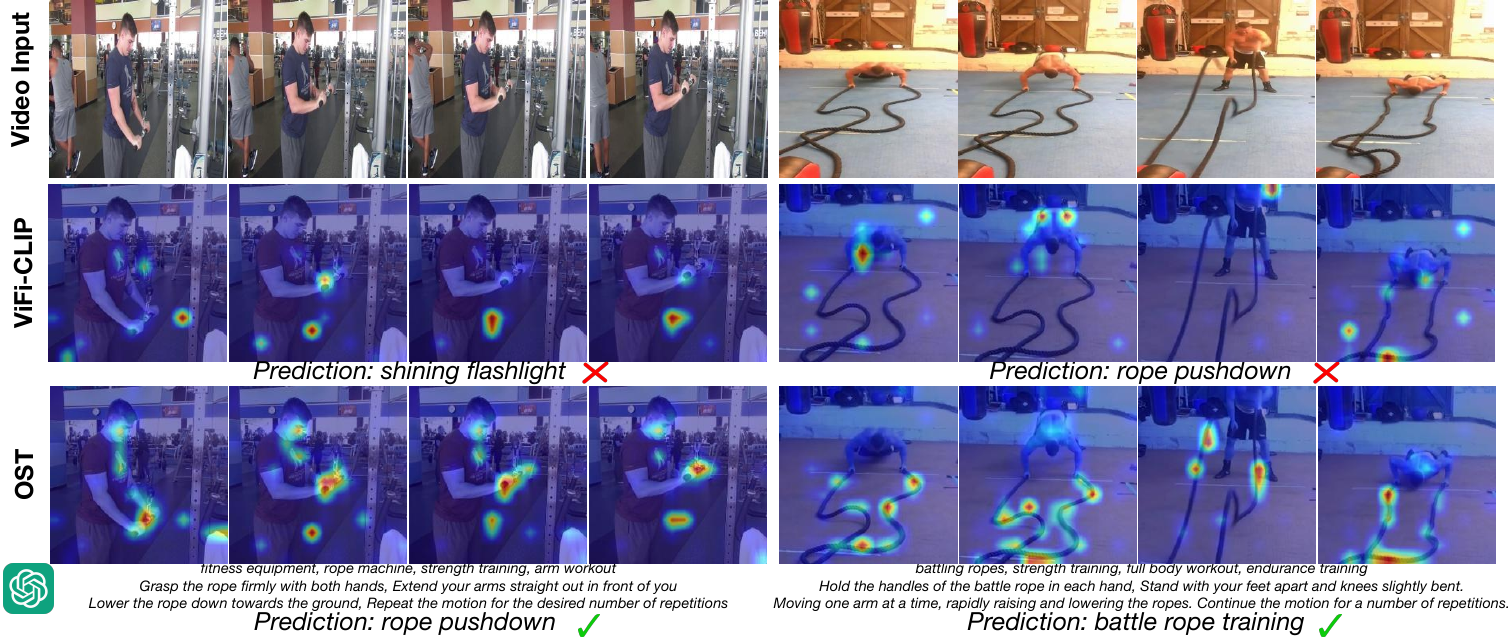}
   \caption{Visualization of cases where our proposed \textit{Spatio-Temporal Descriptors} successfully resolves category mismatch. Relying solely on the category names, ViFi-CLIP misidentifies the equipment as a `\textit{flashlight}’ and misinterprets `\textit{rope pushdown}'. In contrast, aided by \textit{Spatio-Temporal Descriptors}, our \textbf{OST} accurately discerns the action, with a particular focus on temporally significant elements such as the man's hand and the rope.}
   \label{supp: attn_map_rebuttal}
\end{figure*}

We present case studies to illustrate the robustness of our proposed \textbf{OST}, specifically focusing on the transport plan depicted in Fig.~\ref{supp: tp_t_rebuttal} for scenarios where certain action steps are missing, and the attention maps in Fig.~\ref{supp: attn_map_rebuttal} where our \textbf{OST} effectively resolves category mismatches. Detailed analysis is provided within the captions of these figures.
\subsection{Variant of Global Similarity}
Besides the global similarity score computation illustrated in Eq.~\ref{eq:7} in the main paper, an alternative global similarity score can be computed by initially determining the similarity between video representations and descriptor-level embeddings separately, and subsequently averaging these scores to derive the overall global video-descriptor similarity score. Although this approach may appear mathematically analogous to Eq.~\ref{eq:7}, the modified gradient flow during the training process could yield divergent outcomes. As demonstrated in Table~\ref{table: variant_score}, this implementation still exhibits sub-optimal performance in comparison to \textbf{OST}, thereby underscoring the superiority of our proposed method.

\begin{table}[ht] \caption{Study on variants of global similarity score}
\vspace{-1mm}
 \label{table: variant_score}
    \centering
    \setlength{\tabcolsep}{2.0pt}
    \scalebox{0.93}{
\begin{tabular}{l ccc}
    \toprule
    \textbf{Method} & HMDB-51 & UCF-101 & K600\\
    \midrule
    Variant 1 & 53.3 & 76.6 & 69.3\\
    Variant 2 & 52.0 & 76.4 & 69.3\\
    \midrule
    \textbf{OST} & \textbf{54.5} & \textbf{77.9} & \textbf{72.3}\\
    \bottomrule
\end{tabular}}
\vspace{-5mm}
\end{table}

\section{Details of Optimal Descriptor Solver}\label{B}
\subsection{Theoretical Analysis}
In this section, we will provide the theoretical analysis of the existence and unicity of the optimal transport plan $\boldsymbol{P}^{\ast}$ in our proposed \textit{OD Solver}. 

As discussed in Eq.~\ref{eq:2} in the main paper, after obtaining a set of frame-level features $\boldsymbol{V}\in \mathbb{R}^{T\times {d}}$ and descriptor-level embedding for each class $\boldsymbol{D_k^s}\in \mathbb{R}^{N_s\times{d}}$, $\boldsymbol{D_k^t}\in \mathbb{R}^{N_t\times{d}}$. The cost matrix for each class can be defined as:
\begin{equation} \label{eq: supp: 1}
    \boldsymbol{C_k^s} = 1 - cos(\boldsymbol{V}, \boldsymbol{D_k^s}),\quad \boldsymbol{C_k^t} = 1 - cos(\boldsymbol{V}, \boldsymbol{D_k^t}).
\end{equation}
We can define the OT problem in Kantorovich formulation as:
\begin{equation} \label{eq: supp: 2}
\begin{aligned}
    & \boldsymbol{P}^{\ast} = \underset{\boldsymbol{P}\in \mathbb{R}^{T\times N}}{\arg{\min}} \sum^{T}_{i=1}\sum^{N}_{j=1} \boldsymbol{P}_{ij} \boldsymbol{C}_{ij}  \\
    & \textrm{s.t.} \quad \boldsymbol{P} \boldsymbol{e} = \boldsymbol{\mu}, \quad \boldsymbol{P}^{\top}\boldsymbol{e} = \boldsymbol{\nu}.        
\end{aligned}
\end{equation}
However, solving the problem in Eq.~\ref{eq: supp: 2} costs $O(n^3logn)$-complexity, which is time-consuming. By adopting Sinkhorn~\cite{cuturi2013sinkhorn} algorithm, we can define the entropy-regularized OT problem as:
\begin{equation} \label{eq: supp: 3}
\begin{aligned}
    & \boldsymbol{P}^{\ast} = \underset{\boldsymbol{P}\in \mathbb{R}^{T\times N}}{\arg{\min}} \sum^{T}_{i=1}\sum^{N}_{j=1} \boldsymbol{P}_{ij} \boldsymbol{C}_{ij} 
    - \lambda\boldsymbol{H}(\boldsymbol{P})\\
    & \textrm{s.t.} \quad \boldsymbol{P} \boldsymbol{e} = \boldsymbol{\mu}, \quad \boldsymbol{P}^{\top}\boldsymbol{e} = \boldsymbol{\nu}.        
\end{aligned}
\end{equation}
Adding an entropy regularization to the original OT problem makes the optimal regularized transport plan more straightforward. This allows us to calculate the optimal transport distance via Matrix Scaling Algorithms~\cite{sinkhorn1967diagonal}. \\
\noindent\textbf{Lemma 1.} \textit{For $\lambda > 0$, the optimal transport plan $\boldsymbol{P}^{\ast}$ is unique and has the form $\boldsymbol{P}^{\ast} = diag(\boldsymbol{a})\boldsymbol{K}diag(\boldsymbol{b})$, where $\boldsymbol{a}$ and $\boldsymbol{b}$ are two probability vectors of~$\mathbb{R}^{d}$ uniquely defined up to a multiplicative factor and $\boldsymbol{K}=exp(-\boldsymbol{C}/\lambda)$.} \\

\noindent\textit{Proof.} The existence and unicity of $\boldsymbol{P}^{\ast}$ follows from the boundedness of $\boldsymbol{\mu}, \boldsymbol{\nu}$ and the strict convexity of minus the entropy. Consider $\mathcal{L}(P, \alpha, \beta)$ as the Lagrangian of Eq.~\ref{eq: supp: 3}, where $\alpha, \beta$ serve as the dual variables corresponding to the equality constraints in $\boldsymbol{\mu}, \boldsymbol{\nu}$:
\begin{align}
\begin{split}
\mathcal{L}(\boldsymbol{P}, \alpha, \beta) &= \sum_{ij} \left( \frac{1}{\lambda} p_{ij} \log p_{ij} + p_{ij} m_{ij} \right) \\
&\quad+ \alpha^{\top} (\boldsymbol{P} \boldsymbol{e} - \boldsymbol{\mu}) + \beta^{\top} (\boldsymbol{P}^{\top} \boldsymbol{e} - \boldsymbol{\nu}).
\end{split}
\end{align}
For any couple $(i, j)$, if $\left({\partial \mathcal{L}}/{\partial p_{ij}} = 0\right)$, then it follows that $p_{ij} = e^{-{1}/{2}-\lambda_{\alpha_i}} e^{-\lambda m_{ij}} e^{-{1}/{2}-\lambda_{\beta_j}}$. Given that all entries in matrix $\boldsymbol{K}$ are strictly positive, we know from Sinkhorn's work~\cite{sinkhorn1967diagonal} that there is a one-of-a-kind matrix in the form of $diag(\boldsymbol{a})\boldsymbol{K}diag(\boldsymbol{b})$ which fits the constraints given by $\boldsymbol{\mu}, \boldsymbol{\nu}$. Therefore, this matrix is necessarily $\boldsymbol{P}^{\ast}$, and we can calculate it using the Sinkhorn fixed point iteration: 
\begin{equation}
   \boldsymbol{a} \gets \boldsymbol{\mu}/ \boldsymbol{K}\boldsymbol{b}, \quad \boldsymbol{b} \gets \boldsymbol{\nu}/\boldsymbol{K}^{\top}\boldsymbol{a}.
\end{equation}
\qed

\subsection{Pseudo-Code on OD Solver}\label{supp: pseudo codes}
As explained in the paper, our \textit{OD Solver} is effective and simple to implement. In Algorithm~\ref{supp: pseudo codes}, we show the PyTorch style pseudo-code on the implementation of our proposed \textit{Optimal Descriptor Solver}. 
\begin{algorithm*}[h]
	\caption{PyTorch style pseudo-code on Optimal Descriptor Solver}
	\label{supp: pseudo codes}
	\begin{lstlisting}[language=python]
def OptimalDescriptorSolver(video_emb, descriptor_emb):
    A, N, D = descriptor_emb.shape # Get the shape of descriptor embeddings
    B, T, D = video_emb.shape # Get the shape of video embeddings
    sim = torch.einsum('b t d, a n d->t n b a', video_emb, descriptor_emb)  # Compute the similarity
    sim = rearrange(sim, 't n b a->(b a)t n')  # Rearrange dimensions
    cost_mat = 1 - sim  # Calculate the cost matrix
    pp_x = torch.zeros(B*A, T).fill_(1. / T)  # Initialize the horizontal probability vector
    pp_y = torch.zeros(B*A, N).fill_(1. / N)  # Initialize the vertical probability vector
    with torch.no_grad():
        KK = torch.exp( - cost_mat / eps)  # Calculate the cost matrix with exponentiation
        P = Sinkhorn(KK, pp_x, pp_y)  # Apply Sinkhorn algorithm to obtain the optimal transport plan P
        
    # Using optimal transport plan P to obtain logits
    score_ot = torch.sum(P * sim, dim=(1, 2))  # Frobenius inner product
    logits = score_ot.view(B, A)  # Classification logits
    return logits
    
def Sinkhorn(K, u, v):
    r = torch.ones_like(u)  # Initialize r as a tensor of ones with the same shape as u
    c = torch.ones_like(v)  # Initialize c as a tensor of ones with the same shape as v
    thresh = 1e-2  # Threshold to determine convergence in Sinkhorn iterations
    max_iter = 100 # Maximum number of Sinkhorn iterations
    # Sinkhorn iteration
    for i in range(max_iter):  # Iterate up to the maximum number of iterations
        r0 = r # Save the previous iteration's r
        r = u / torch.matmul(K, c.unsqueeze(-1)).squeeze(-1)  # Update r
        c = v / torch.matmul(K.permute(0, 2, 1), r.unsqueeze(-1)).squeeze(-1)  # Update c
        err = (r - r0).abs().mean()  # Calculate the mean absolute change in iterations
        if err.item() < thresh:  # If the change is below the threshold, stop iterating
            break
        P = torch.matmul(r.unsqueeze(-1), c.unsqueeze(-2)) * K  # Obtain the final transport plan P
    return P
    \end{lstlisting}
\end{algorithm*}

\section{Implementation Details}\label{C}
\subsection{Dataset Details}\label{supp: dataset details}
We provide 6 video benchmarks used in our empirical studies:

\noindent\textbf{Kinetic-400}~\cite{quovadis} is a large-scale video dataset consisting of 10-second video clips collected from YouTube. 240,000 training videos and 20,000 validation videos in 400 different action categories. 

\noindent\textbf{Kinetic-600}~\cite{k600} is an extension of Kinetics-400, consisting of approximarely 480,000 videos from 600 action categories. The videos are divided into 390,000 for training, 30,000 for validation, and 60,000 for testing. We mainly use its validation set for zero-shot evaluation.

\noindent\textbf{UCF-101}~\cite{soomro2012ucf101} is a video recognition dataset for realistic actions, collected from YouTube, including 13,320 video clips with 101 action categories in total. There are three splits of the training and testing data.

\noindent\textbf{HMDB-51}~\cite{kuehne2011hmdb} is a relatively small video dataset compared to Kinetics and UCF-101. It has around 7,000 videos with 51 classes. HMDB-51 has three splits of the training and testing data.

\noindent\textbf{Something-Something V2}~\cite{goyal2017something} is a challenging temporal-heavy dataset which contains 220,000 video clips across 174 fine-grained classes.

\noindent\textbf{ActivityNet}~\cite{caba2015activitynet} We use the ActivityNet-v1.3 in our experiments. ActivityNet is a large-scale untrimmed video benchmark, containing 19,994 untrimmed videos of 5 to 10 minutes from 200 activity categories.

\subsection{Implementation Details}\label{supp: implementation details}
\noindent\textbf{Zero-shot Experiments.}
We mainly follow the zero-shot setting in~\cite{vificlip, ni2022expanding}. We tune both the visual and textual encoder of a CLIP ViT-B/16 with 32 frames on Kinetics-400 for 10 epochs. The batch size is set as 256 and single-view inference is adopted during validation. We set the hyperparameters in the Sinkhorn algorithm~\cite{cuturi2013sinkhorn} as $\lambda=0.1$. We adopt the AdamW optimizer paired with a $8\times10^{-6}$ initial learning rate with the CosineAnnealing learning rate schedule. Following~\cite{lin2023match, ilharco2022patching, wortsman2022robust}, we perform a linear weight-space ensembling between the original CLIP model and the finetuned model with a ratio of $0.2$.

We apply the following evaluation protocols in our zero-shot experiments: For UCF-101 and HMDB-51, the prediction is conducted on three official splits of the test data. We report average Top-1 accuracy and standard deviation. For Kinetics-600, following~\cite{chen2021elaborative}, the 220 new categories outside Kinetics-400 are used for evaluation. We use the three splits provided by~\cite{chen2021elaborative} and sample 160 categories for evaluation from the 220 categories in Kinetics-600 for each split. We report average Top-1 and Top-5 accuracy and standard deviation.

\noindent\textbf{Few-shot Experiments.} 
For the few-shot setting, we utilize CLIP ViT-B/16 as  We adopt the few-shot split from~\cite{vificlip, ni2022expanding} that randomly samples 2, 4, 8, and 16 videos from each class on UCF-101, HMDB-51, and Something-Something V2 for constructing the training set. For evaluation, we use the first split of the test set on UCF-101, HMDB-51, and Something-Something V2. We utilize 32 frames during training and validation. Top-1 accuracy with single-view inference is reported. We set the batch size as 64 and train for 50 epochs in few-shot experiments.

\noindent\textbf{Fully-supervised Experiments.}
For fully-supervised studies, we base our approach on Text4Vis~\cite{text4vis} to conduct experiments in frozen text settings and keep the hyperparameters and data augmentations consistent with the baseline. We vary CLIP ViT-B/32, and ViT-B/16 as encoder and train with 8, and 16 frames, respectively. We report Top-1 accuracy using single-view inference.

\noindent\textbf{Data Augmentation Recipe.}
For a fair comparison, we largely follow the data augmentations in ViFi-CLIP~\cite{vificlip} for zero-shot and few-shot experiments and follow the recipe in Text4Vis~\cite{text4vis} for fully-supervised experiments.  The details for our data augmentation recipe are shown in Table~\ref{supp: augmentation}.
\begin{table}[ht] \caption{Data augmentation recipe for video recognition.}
\vspace{-2mm}
 \label{supp: augmentation}
    \centering
    \setlength{\tabcolsep}{2.0pt}
    \scalebox{0.93}{
\begin{tabular}{l c c}
  \toprule
  Setting   & Zero/Few-shot & Fully-supervised \\
 \hline 
 \rowcolor{gray!20}\multicolumn{3}{l}{\textit{Augmentation}} \\ 
RandomFlip & 0.5 & 0.5 \\
Crop & \textit{MultiScaleCrop} & \textit{RandomSizedCrop} \\
ColorJitter  & 0.8 & 0 \\
GrayScale & 0.2 & 0.2 \\
Label smoothing & 0 & 0  \\
Mixup & 0 & 0 \\
Cutmix & 0 & 0 \\
\bottomrule
\end{tabular}}
\vspace{-5mm}
\end{table}

\noindent\textbf{Training and Testing.}
We employ the identical alignment mechanism throughout both the training and testing phases. The only difference lies in the application of contrastive-style operations during training, where logits are obtained exclusively from descriptors within the current mini-batch. During testing, classification scores are calculated against descriptors from all classes.

\section{Demonstration of Prompts and Descriptors}\label{D}
\subsection{Prompting the Language Model}\label{supp: prompts}
\begin{figure*}
  \centering
   \includegraphics[width=0.73\linewidth]{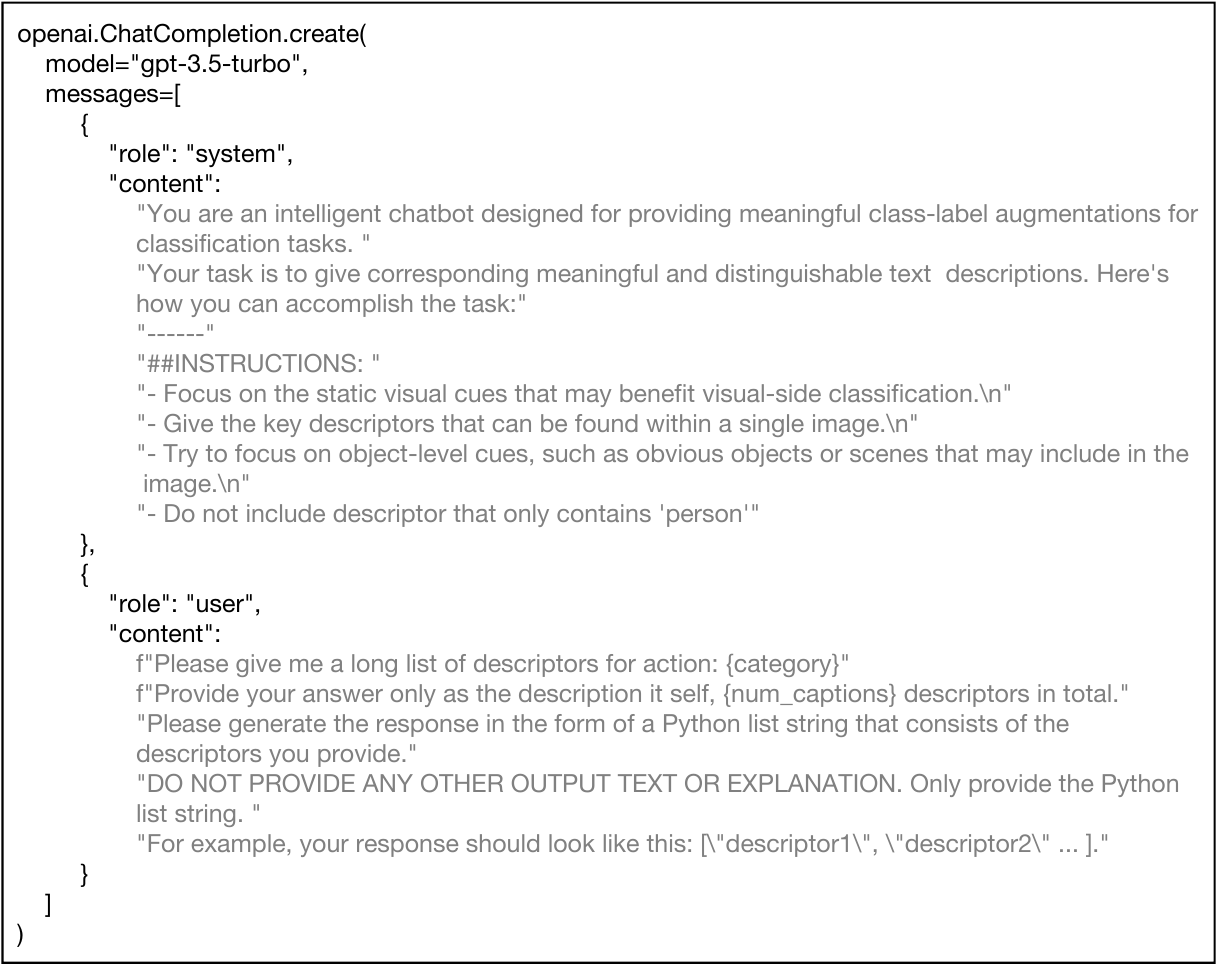}
   \caption{Prompt for generating \textit{Spatio Descriptors}. The generated \textit{Spatio Descriptors} are intended to capture static visual elements that can be discerned from a single image, such as environments and objects. So we prompt the LLM to prioritize and interpret object-level cues.}
   \label{fig: prompt_spatio}
\end{figure*}

\begin{figure*}
  \centering
   \includegraphics[width=0.73\linewidth]{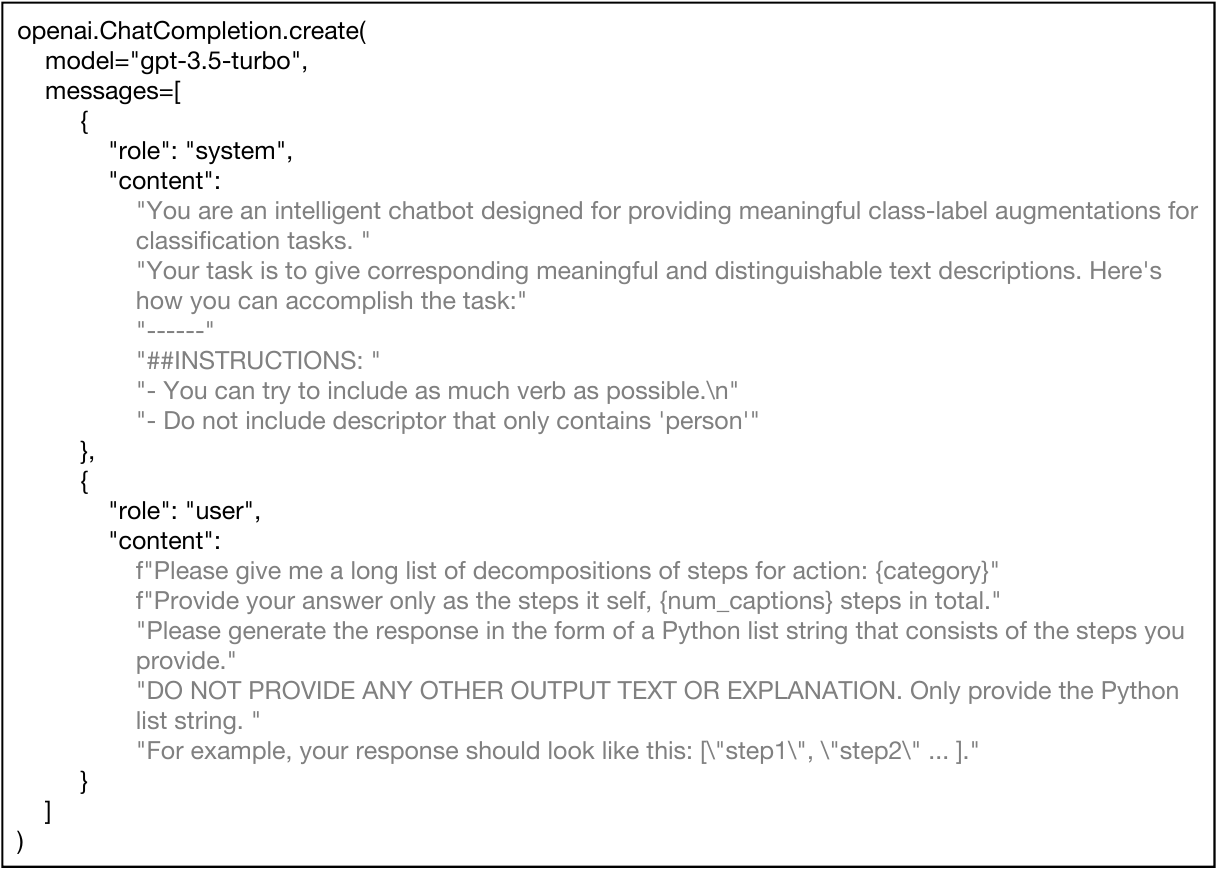}
   \caption{Prompt for generating \textit{Temporal Descriptors}. For \textit{Temporal Descriptors}, our aim is to decompose the action classes in a step-by-step manner, detailing how an action progresses over time. To enhance the adapted model's capacity to learn action verbs during the training phase, we prompt the LLM to include a comprehensive range of verbs.}
   \label{fig: prompt_temporal}
\end{figure*}

We provide our prompts for generating \textit{Spatio-Temporal Descriptors} in Fig.~\ref{fig: prompt_spatio} and Fig.~\ref{fig: prompt_temporal}, respectively. We provide details in the figure captions.

\subsection{Additional Examples of Spatio-Temporal Descriptors}\label{supp: descriptors}
In this section, we provide additional examples of the \textit{Spatio-Temporal Descriptors}. \\
\noindent\textbf{Descriptors for action category ``Adjusting Glasses":}

{\scriptsize\noindent\textit{Spatio Descriptor}:}
\texttt{
    {\begin{enumerate}
        \scriptsize
        \setlength{\itemsep}{0pt}%
        \setlength{\parskip}{0pt}%
        \item person wearing glasses
        \item hand adjusting glasses
        \item glasses sliding on face
        \item fingers pushing up glasses
    \end{enumerate}}
}

{\scriptsize\noindent\textit{Temporal Descriptor}:}
\texttt{
    {\begin{enumerate}
        \scriptsize
        \setlength{\itemsep}{0pt}%
        \setlength{\parskip}{0pt}%
        \item Push the glasses up the bridge of your nose
        \item Align the temples with your ears
        \item Adjust the nose pads for comfort
        \item Ensure that the glasses rest comfortably on your face
    \end{enumerate}}
}

\noindent\textbf{Descriptors for action category ``Assembling Bicycle":}

{\scriptsize\noindent\textit{Spatio Descriptor}:}
\texttt{
    {\begin{enumerate}
        \scriptsize
        \setlength{\itemsep}{0pt}%
        \setlength{\parskip}{0pt}%
        \item Bicycle frame
        \item Handlebars
        \item Wheels
        \item Pedals
    \end{enumerate}}
}

{\scriptsize\noindent\textit{Temporal Descriptor}:}
\texttt{
    {\begin{enumerate}
        \scriptsize
        \setlength{\itemsep}{0pt}%
        \setlength{\parskip}{0pt}%
        \item Attach the front wheel to the bicycle frame using a wrench and follow the specified torque setting.
        \item Secure the handlebars onto the front fork by tightening the stem bolts with an Allen wrench.
        \item Install the pedals onto the crank arms by screwing them in clockwise.
        \item Adjust the seat height to the desired position and tighten the seat clamp to secure it.
    \end{enumerate}}
}

\noindent\textbf{Descriptors for action category ``Building Sandcastle":}

{\scriptsize\noindent\textit{Spatio Descriptor}:}
\texttt{
    {\begin{enumerate}
        \scriptsize
        \setlength{\itemsep}{0pt}%
        \setlength{\parskip}{0pt}%
        \item beach
        \item sand
        \item castle
        \item bucket
    \end{enumerate}}
}

{\scriptsize\noindent\textit{Temporal Descriptor}:}
\texttt{
    {\begin{enumerate}
        \scriptsize
        \setlength{\itemsep}{0pt}%
        \setlength{\parskip}{0pt}%
        \item Dig a shallow hole in the sand for the base
        \item Fill the hole with wet sand and pack it down firmly
        \item Create a large mound of sand on top of the base
        \item Use your hands or tools to shape the sand into walls and towers
    \end{enumerate}}
}

\noindent\textbf{Descriptors for action category ``Opening Wine Bottle":}

{\scriptsize\noindent\textit{Spatio Descriptor}:}
\texttt{
    {\begin{enumerate}
        \scriptsize
        \setlength{\itemsep}{0pt}%
        \setlength{\parskip}{0pt}%
        \item wine bottle
        \item corkscrew
        \item uncorking
        \item pouring
    \end{enumerate}}
}

{\scriptsize\noindent\textit{Temporal Descriptor}:}
\texttt{
    {\begin{enumerate}
        \scriptsize
        \setlength{\itemsep}{0pt}%
        \setlength{\parskip}{0pt}%
        \item Hold the wine bottle firmly
        \item Remove the foil or plastic covering from the top of the bottle
        \item Insert the corkscrew into the center of the cork
        \item Twist the corkscrew counterclockwise to remove the cork
    \end{enumerate}}
}

\noindent\textbf{Descriptors for action category ``Planing Wood":}

{\scriptsize\noindent\textit{Spatio Descriptor}:}
\texttt{
    {\begin{enumerate}
        \scriptsize
        \setlength{\itemsep}{0pt}%
        \setlength{\parskip}{0pt}%
        \item wood
        \item sawdust
        \item saw
        \item workbench
    \end{enumerate}}
}

{\scriptsize\noindent\textit{Temporal Descriptor}:}
\texttt{
    {\begin{enumerate}
        \scriptsize
        \setlength{\itemsep}{0pt}%
        \setlength{\parskip}{0pt}%
        \item Measure and mark the dimensions of the wood piece
        \item Cut the wood according to the marked measurements
        \item Smooth the edges of the cut wood using sandpaper
        \item Apply a coat of varnish or paint to protect and enhance the appearance of the wood
    \end{enumerate}}
}





\section{Broader Impact and Limitation}\label{H}
\textbf{OST} represents an effective way to utilize external knowledge to adapt pre-trained visual-language models for general video recognition. Our approach can benefit zero-shot, few-shot, and fully-supervised video recognition with no modification to the model architecture and minor additional computational costs. Furthermore, the proposed \textit{Spatio-Temporal Descriptor} can greatly reduce the semantic similarity of action categories. \textbf{The employment of LLMs to generate corresponding descriptors can be readily extended to various unseen action categories, allowing the open-vocabulary understanding of actions in the wild.}

However, the quality of descriptors directly connects to the final performance. The process of generating descriptors highly depends on the knowledge learned by the LLM, which is only partially controllable by varying the prompts. Additionally, our findings suggest that the informational needs for describing actions differ across various categories. Relying solely on four Spatio-Temporal Descriptors might not be ideal for every category. An adaptive approach, where the number of descriptors is tailored to each category, would likely be more effective.


\end{document}